%% file: main.tex
\documentclass[sigconf]{acmart}

\AtBeginDocument{}

\setcopyright{acmlicensed}
\copyrightyear{2027}
\acmYear{2027}
\acmDOI{}
\acmConference[KDD '27]{The 33rd ACM SIGKDD Conference on Knowledge Discovery and Data Mining}{August 1--5, 2027}{San Jose, CA, USA}
\acmISBN{}
\usepackage{booktabs}
\usepackage{graphicx}
\usepackage{fontawesome5}
\usepackage{siunitx}
\usepackage{makecell}
\usepackage{tabularx}
\usepackage{longtable}
\usepackage{multirow}
\usepackage{placeins}
\usepackage{tikz}
\usetikzlibrary{arrows.meta,positioning,fit,calc}

\begin{document}

\title{KVDiagnosis: A Diagnostic Benchmark for KV-Cache Compression in Long-Context Language Models}

\author{Chen Qiu}
\affiliation{%
  \institution{KAUST}
  \city{Thuwal}
  \country{Saudi Arabia}}
\email{chen.qiu@kaust.edu.sa}

\author{Ziwu Liu}
\affiliation{%
  \institution{KAUST}
  \city{Thuwal}
  \country{Saudi Arabia}}
\email{ziwu.liu@kaust.edu.sa}

\author{Chao Fei}
\affiliation{%
  \institution{KAUST}
  \city{Thuwal}
  \country{Saudi Arabia}}
\email{chao.fei@kaust.edu.sa}

\author{Guozhong Li}
\affiliation{%
  \institution{KAUST}
  \city{Thuwal}
  \country{Saudi Arabia}}
\email{guozhong.li@kaust.edu.sa}

\author{Panos Kalnis}
\affiliation{%
  \institution{KAUST}
  \city{Thuwal}
  \country{Saudi Arabia}}
\email{panos.kalnis@kaust.edu.sa}

\renewcommand{\shortauthors}{Qiu et al.}

\input{sections/abstract}

% \keywords{KV-cache compression, long-context language models, diagnostic benchmarking, paired evaluation, failure profiling}

\maketitle

\input{sections/introduction}
\input{sections/related}
\input{sections/taxonomy}
\input{sections/metirc}
\input{sections/experiment}
\input{sections/conclusion}

\newpage
\vfuzz=0.1pt
\nobalance
\bibliographystyle{ACM-Reference-Format}
\bibliography{references}

\clearpage
\input{sections/appendix}

\end{document}

%% file: sections/abstract.tex
\begin{abstract}
KV-cache compression reduces long-context memory, but aggregate task scores reveal neither which correct executions fail nor why. We present \textsc{KVDiagnosis}, a diagnostic dataset and benchmark with three contributions. First, a 25-method taxonomy groups methods into five mechanism families and links them to eight verified implementations and their valid diagnostic measurements. Second, for every supported method--setting, we evaluate all sources in each fixed split against a per-source FullCache control before selecting FullCache-correct/compressed-wrong (C$\rightarrow$W) rows separately for each method--setting, so no compressor defines another's test set. Third, a common record format links paired outputs and run metadata to cache, likelihood, attention, and decoding measurements with explicit applicability states. On Qwen3-8B, four evidence-aware workloads yield \num{59800} supported compressed runs over \num{2600} sources and \num{12520} C$\rightarrow$W rows. Under fixed diagnostic rules, 63.2\% have low or partial measured/projected coverage. Only 19 rows (0.2\%) combine high measured/projected coverage with strong likelihood drift; another 2,126 (17.0\%) preserve structural position addressability, for which representation fidelity remains unknown, while showing the same drift. Against C$\rightarrow$C success controls, all ten diagnostics separate failed from successful compression (stratified AUROC 0.684--0.871). Among 96 reproducible low-EAR failures, a controlled 4$\times$ evidence-attention boost repairs 29.2\%, versus 6.3\% under a count-matched sham intervention and 3.3\% degradation on matched C$\rightarrow$C controls. Code and data are available at \href{https://github.com/ChosenQC/KVDiagnosis}{\textbf{\faGithub\ KVDiagnosis}}.
\end{abstract}

%% file: sections/introduction.tex
% !TEX root = ../main.tex
\section{Introduction}\label{sec:introduction}

Long-context language models make document-scale question answering, retrieval, and reasoning possible, but their usable context is constrained by the state carried through autoregressive decoding. Every additional context token contributes a key and value at every layer, so KV-cache memory grows with sequence length and concurrent requests. 
Compression directly targets this state and is therefore an important route to practical long-context inference. It also creates an evaluation problem: when a previously correct answer breaks, a memory-footprint statistic and final task score do not identify what changed.\looseness=-1

KV-cache compressors transform different objects. Some evict token positions, others redistribute a fixed budget across layers or heads, and still others prune channels, quantize tensors, or retain semantic units~\cite{xiao2024efficient,li_snapkv_2024,feng2024adakv,xu2024think,liu2024kivi,liu2025chunkkv}. Each mechanism changes a different cache object and exposes different measurements. A label such as ``50\%'' may mean half the positions for one method, fewer key channels for another, and lower numerical precision for a third. A fair benchmark must state what each setting means and use only diagnostics valid for its transformed object.\looseness=-1

Figure~\ref{fig:taxonomy} makes this variation explicit by organizing 25 representative methods into five primary mechanism families. Each method is placed by its main compression decision and transformed cache object; secondary traits may cross family boundaries. The map provides survey coverage, while boldface marks eight verified implementations selected for evaluation across all five families. This separation keeps the literature coverage broader than the claims supported by our experiments.\looseness=-1

\input{sections/taxonomy_figure}

Existing resources address complementary parts of this problem. Long-context benchmarks provide broad or controlled tasks~\cite{bai2023longbench,hsieh2024ruler,li2024scbench}, while surveys and cross-method studies organize compressors and report quality--efficiency trade-offs~\cite{kv_cache_management_survey,shi2024keepcost,yuan2024kvbenchmark}. Gao et al. additionally identify per-sample performance drops and release a thresholded negative-sample benchmark, while Chen et al. characterize instruction-level failures and test changes to eviction methods~\cite{rethinking_kv_cache_compression,chen2026pitfalls}. These resources establish that aggregate scores can hide failures on individual examples. Yet no public resource we identified jointly provides (i) complete per-source results with FullCache controls; (ii) method--setting-specific FullCache-correct to compressed-wrong (C$\rightarrow$W) records selected only after all source-level runs are complete; and (iii) valid cache, likelihood, attention, and decoding traces for those records. To our knowledge, no benchmark connects all three in one reusable diagnostic dataset.\looseness=-1

This missing connection matters in two ways. First, the same final answer error can arise because a compressor deletes answer-supporting context, damages the representations stored in retained entries, weakens generation-time access, or changes the result during decoding or scoring. These causes require different fixes: higher precision cannot recover evicted evidence, and retaining more positions cannot repair a representation error. Second, failure sets are method-specific. Evaluating one method only on failures found for another changes the test distribution and can hide distinctive cases, even when their aggregate scores are comparable. A ranking alone therefore cannot explain a failure, select a repair, or show that two compressors are interchangeable.\looseness=-1

We introduce \textsc{KVDiagnosis}, a paired diagnostic benchmark designed around this distinction. FullCache is executed once per source; every supported method--setting cell then covers all sources in the fixed benchmark split with the same model, prompt, tokenizer, decoder, and scorer. After completing the run matrix, the benchmark extracts each cell's C$\rightarrow$W rows and attaches every applicable diagnostic. Unsupported settings, execution errors, and inapplicable traces retain distinct status codes rather than disappearing from the accounting. A method's own failure set shows what it breaks, while intersections are used only for matched comparisons. This preserves scores and denominators over all sources without letting failure selection redefine the evaluated source set.\looseness=-1

On Qwen3-8B across four evidence-aware workloads~\cite{qwen2025qwen3}, the benchmark contains \num{59800} supported compressed runs over \num{2600} sources and \num{12520} method--setting C$\rightarrow$W rows. Under fixed diagnostic rules, 63.2\% have low or partial measured/projected support coverage. High measured/projected coverage with strong likelihood drift is rare (19 rows), whereas 2,126 rows show drift under structural position addressability, which does not measure representation fidelity.
Methods with comparable aggregate quality can nevertheless fail on different source sets.\looseness=-1

\noindent\textbf{Contributions.} This paper makes the following contributions.
\begin{itemize}\setlength{\itemsep}{1pt}\setlength{\topsep}{2pt}
    \item \textbf{Mechanism coverage.} We synthesize a representative 25-method taxonomy and a clear mapping from five mechanism families to eight verified implementations and the diagnostics each supports.
    \item \textbf{Evaluation on all sources.} We pair every source in each fixed split with FullCache, select failures separately for each method--setting, and link outputs and run metadata to valid cache, likelihood, attention, and decoding measurements without changing the evaluated source set.
    \item \textbf{Paired empirical resource.} We construct \num{59800} supported compressed runs and \num{12520} failure rows across four workloads, validate the diagnostics against C$\rightarrow$C controls, and test low-EAR failures with a targeted repair.
\end{itemize}

%% file: sections/taxonomy_figure.tex
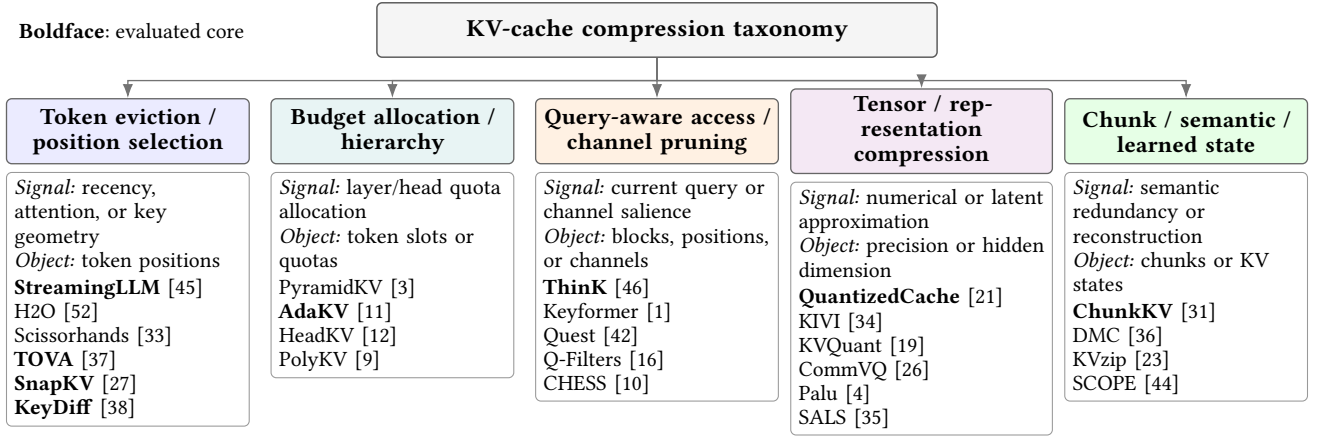
\begin{figure*}[t!]
\centering
\begin{tikzpicture}[
    font=\fontsize{8.2pt}{9.1pt}\selectfont,
    >=Latex,
    root/.style={
        draw=black!65,
        rounded corners=2pt,
        line width=0.55pt,
        fill=black!4,
        text width=7.2cm,
        minimum height=0.72cm,
        inner ysep=2.5pt,
        align=center,
        font=\fontsize{9.2pt}{10pt}\selectfont\bfseries
    },
    family/.style={
        draw=black!60,
        rounded corners=2pt,
        line width=0.50pt,
        text width=3.00cm,
        minimum height=0.86cm,
        inner ysep=2.5pt,
        align=center,
        font=\fontsize{8.7pt}{9.5pt}\selectfont\bfseries
    },
    detail/.style={
        draw=black!40,
        rounded corners=2pt,
        line width=0.40pt,
        fill=white,
        text width=3.00cm,
        inner xsep=3pt,
        inner ysep=2.5pt,
        align=left
    },
    edge/.style={->, line width=0.50pt, draw=black!55}
]

\node[root] (root) at (0,0)
  {KV-cache compression taxonomy};
\node[anchor=west, font=\fontsize{8.2pt}{9.1pt}\selectfont] at (-8.55,0)
  {\textbf{Boldface}: evaluated core};

\node[family, fill=blue!8] (evict) at (-7.00,-1.34)
  {Token eviction /\\position selection};
\node[family, fill=teal!9] (budget) at (-3.50,-1.34)
  {Budget allocation /\\hierarchy};
\node[family, fill=orange!11] (query) at (0,-1.34)
  {Query-aware access /\\channel pruning};
\node[family, fill=violet!9, text width=3.25cm] (quant) at (3.50,-1.34)
  {Tensor / representation\\compression};
\node[family, fill=green!10] (state) at (7.00,-1.34)
  {Chunk / semantic /\\learned state};

\node[detail, below=0.10cm of evict] (evictdetail) {
  \textit{Signal:} recency, \mbox{attention}, or key \mbox{geometry}\\
  \textit{Object:} token positions\\[1pt]
  \textbf{StreamingLLM}~\cite{xiao2024efficient}\\
  H2O~\cite{zhang_h_2o_2023}\\
  Scissorhands~\cite{liu2023scissorhands}\\
  \textbf{TOVA}~\cite{oren2024transformers}\\
  \textbf{SnapKV}~\cite{li_snapkv_2024}\\
  \textbf{KeyDiff}~\cite{park2025kediff}
};

\node[detail, below=0.10cm of budget] (budgetdetail) {
  \textit{Signal:} layer/head quota allocation\\
  \textit{Object:} token slots or quotas\\[1pt]
  PyramidKV~\cite{cai2024pyramidkv}\\
  \textbf{AdaKV}~\cite{feng2024adakv}\\
  HeadKV~\cite{fu2024headkv}\\
  PolyKV~\cite{fei2026polykv}
};

\node[detail, below=0.10cm of query] (querydetail) {
  \textit{Signal:} current query or channel salience\\
  \textit{Object:} blocks, positions, or channels\\[1pt]
  \textbf{ThinK}~\cite{xu2024think}\\
  Keyformer~\cite{adnan2024keyformer}\\
  Quest~\cite{tang2024quest}\\
  Q-Filters~\cite{godey2025qfilters}\\
  CHESS~\cite{fei2026chess}
};

\node[detail, text width=3.25cm, below=0.10cm of quant] (quantdetail) {
  \textit{Signal:} numerical or latent approximation\\
  \textit{Object:} precision or hidden dimension\\[1pt]
  \textbf{QuantizedCache}~\cite{huggingface_quantizedcache}\\
  KIVI~\cite{liu2024kivi}\\
  KVQuant~\cite{hooper2024kvquant}\\
  CommVQ~\cite{li2025commvq}\\
  Palu~\cite{chang2025palu}\\
  SALS~\cite{mu2025sals}
};

\node[detail, below=0.10cm of state] (statedetail) {
  \textit{Signal:} \mbox{semantic} \mbox{redundancy} or \mbox{reconstruction}\\
  \textit{Object:} chunks or KV states\\[1pt]
  \textbf{ChunkKV}~\cite{liu2025chunkkv}\\
  DMC~\cite{nawrot2024dynamic}\\
  KVzip~\cite{kim2026kvzip}\\
  SCOPE~\cite{wu2025scope}
};

\draw[edge] (root.south) -- ++(0,-0.30) -| (evict.north);
\draw[edge] (root.south) -- ++(0,-0.30) -| (budget.north);
\draw[edge] (root.south) -- (query.north);
\draw[edge] (root.south) -- ++(0,-0.30) -| (quant.north);
\draw[edge] (root.south) -- ++(0,-0.30) -| (state.north);

\end{tikzpicture}
\caption{KV-cache compression taxonomy and evaluated subset. Each method is assigned by its primary compression decision and transformed object; bold names denote the eight verified implementations. Table~\ref{tab:method_landscape} in the Appendix provides the detailed method-to-diagnostic mapping.}
\Description{A taxonomy of 25 KV-cache compression methods in five families. Bold names identify the eight methods evaluated in the benchmark, while regular names identify literature-only methods.}
\label{fig:taxonomy}
\end{figure*}

%% file: sections/related.tex
% !TEX root = ../main.tex
\section{Related Work}\label{sec:related_work}

We organize related work into two groups: compression mechanisms that define what a method changes, and long-context resources and systems that define what existing evaluations report. Table~\ref{tab:benchmark_positioning} previews the resulting benchmark-level distinction.

\begin{table*}[t]
\centering
\fontsize{8.1pt}{8.8pt}\selectfont
\setlength{\tabcolsep}{2.1pt}
\renewcommand{\arraystretch}{0.88}
\caption{Released support for attributing and diagnosing KV-cache compression failures (C$\rightarrow$W: correct under FullCache, wrong after compression). Grouped rows report the union across cited resources.}
\label{tab:benchmark_positioning}
\begin{tabular}{@{}>{\raggedright\arraybackslash}p{0.22\textwidth}
  >{\centering\arraybackslash}p{0.235\textwidth}
  >{\centering\arraybackslash}p{0.13\textwidth}
  >{\centering\arraybackslash}p{0.15\textwidth}
  >{\raggedright\arraybackslash}p{0.225\textwidth}@{}}
\toprule
\textbf{Resource} & \textbf{Released outputs} & \textbf{Matched FullCache baseline} & \textbf{Failure cohort} & \textbf{Diagnostic measurements} \\
\midrule
Long-context task suites~\cite{bai2023longbench,hsieh2024ruler} & Task-level aggregates & None & None & None \\
Aggregate KV evaluations~\cite{li2024scbench,yuan2024kvbenchmark} & Task--method aggregates & Aggregate only & None & Study-level attention/behavior \\
Negative-sample benchmark~\cite{rethinking_kv_cache_compression} & Selected source-level failures & Per source & Score-loss threshold & Outcome degradation \\
Failure-mode study~\cite{chen2026pitfalls} & Study-specific source/instruction cases & Per source & Study-defined & Eviction bias; instruction behavior \\
\textbf{\textsc{KVDiagnosis} (Ours)} & Complete source--method--setting matrix & Matched per source & C$\rightarrow$W after matrix completion & Cache retention/coverage; paired logits, attention, decoding \\
\bottomrule
\end{tabular}
\end{table*}

\subsection{KV-Cache Compression Landscape}
Figure~\ref{fig:taxonomy} organizes 25 representative methods by their primary compression decision and transformed cache object. Position methods use recency, attention, or key geometry~\cite{xiao2024efficient,zhang_h_2o_2023,liu2023scissorhands,oren2024transformers,li_snapkv_2024,park2025kediff}. Allocation methods redistribute slots across layers or heads, including recent head-level and heterogeneous layer-wise policies~\cite{cai2024pyramidkv,feng2024adakv,fu2024headkv,fei2026polykv}. Query-aware access and channel-pruning methods select pages, tokens, or channels at access time~\cite{xu2024think,adnan2024keyformer,tang2024quest,godey2025qfilters,fei2026chess}. Tensor/representation methods range from scalar and vector quantization to low-rank latent caches~\cite{huggingface_quantizedcache,liu2024kivi,hooper2024kvquant,li2025commvq,chang2025palu,mu2025sals}; chunk, semantic, and learned-state methods compress larger units~\cite{liu2025chunkkv,nawrot2024dynamic,kim2026kvzip,wu2025scope}.

Other designs not included in the main comparison classify attention heads or estimate expected attention~\cite{ge2024fastgen,devoto2025expectedattention}; mix precision, product quantization, or coupled low-bit codes~\cite{yang2024mikv,zhang2024pqcache,he2024zipcache,zhang2024coupledquant}; and merge states across tokens and layers~\cite{liu2024minicache,lancucki2026inference,gelberg2026kvcat}. ART instead skips negligible KV blocks during attention and complements, rather than implements, a retained-cache policy~\cite{qiu2026art}. Reviews summarize this rapidly changing space~\cite{kv_cache_management_survey,shi2024keepcost}; related studies compare long-context approaches, deployed-system behavior, or data-dependent compressibility~\cite{yuan2024kvbenchmark,rethinking_kv_cache_compression,chen2026kvcore}. Our representative taxonomy defines transformed objects and valid measurements.

The same measurements do not apply unchanged to every family. Position retention is measured for token eviction, projected from chunks, \texttt{N/A} when positions are only structurally addressable, and potentially undefined for learned state. Cross-family diagnostics therefore require mechanism-specific interpretation. The detailed Appendix table (Table~\ref{tab:method_landscape}) mirrors Figure~\ref{fig:taxonomy}; its \textbf{Evaluated} entries identify the eight implementations spanning all five mechanisms.

\subsection{Long-Context and KV-Cache Benchmarks}
LongBench provides broad long-document tasks, while RULER offers controlled retrieval, tracking, and aggregation at configurable lengths~\cite{bai2023longbench,hsieh2024ruler}. Lost in the Middle shows that evidence position affects long-context performance~\cite{liu2024lostmiddle}. Qasper and HotpotQA provide evidence-annotated questions~\cite{dasigi2021qasper,yang2018hotpotqa}; SCBench studies cache generation, compression, retrieval, and loading over multi-turn workloads~\cite{li2024scbench}. No reviewed public resource combines complete per-source method--setting results, cell-specific C$\rightarrow$W rows, and valid cache and predictive-distribution measurements.\looseness=-1

CacheGen compresses KV tensors for streaming~\cite{liu2024cachegen}; serving methods optimize memory movement and execution, but neither replaces paired correctness analysis. Recent benchmarks make reusable data, evaluation, and release requirements explicit through production workloads, standardized frameworks, or separated capability dimensions~\cite{wang2025burstgpt,gu2025vflairllm,jiang2025hibench}. \textsc{KVDiagnosis} applies this release-oriented design to compression failures. Matched pairs hold model, prompt, tokenizer, decoding, and scorer fixed; complete coverage requires every planned cell or explicit N/A.\looseness=-1

%% file: sections/taxonomy.tex
% !TEX root = ../main.tex
\section{Benchmark Design}\label{sec:taxonomy}

We define the eligible mechanisms, construct the complete paired run matrix, and specify when each diagnostic applies. We separate overall quality from failure analysis while retaining the source and configuration details needed for reproducible reuse.

\subsection{Scope and Evaluation Units}
The scope follows three design constraints. First, the benchmark covers inference-time transformations that change the stored KV representation or the entries available to one attention operation while model parameters remain fixed. Architectural KV-head reduction, paging, offloading, scheduling, and cross-request reuse are related mechanisms rather than benchmark subjects. Second, the evaluated methods must span distinct transformed objects in the survey taxonomy. Third, an implementation is included only after tests confirm that it runs the intended compression method and produces every expected result.\looseness=-1

The resulting eight evaluated methods span position selection, head-wise allocation, channel pruning, chunk retention, and tensor precision. Figure~\ref{fig:taxonomy} shows this family coverage; Table~\ref{tab:benchmark_scope} lists each implementation's transformed unit and valid cache measurements. Seven methods use \texttt{kvpress} 0.5.3; QuantizedCache uses the Transformers/HQQ cache implementation~\cite{huggingface_quantizedcache}.\looseness=-1

Four record types prevent ambiguous claims. A \emph{source} is one task prompt. A \emph{run record} pairs a source with one method and configured setting, including settings that the implementation does not support. A \emph{supported run} completes compressed inference. A \emph{failure row} is a supported run whose paired FullCache answer is correct and compressed answer is wrong. Because one source can yield several failures, run-, row-, and unique-source counts differ. Figure~\ref{fig:overview} summarizes how the complete paired run matrix becomes a diagnostic dataset without changing the evaluated source set.\looseness=-1

\begin{table*}[t]
\centering
\small
\setlength{\tabcolsep}{2.5pt}
\caption{Fixed benchmark scope. Top: source splits and evidence metadata. Bottom: valid cache measurements for the eight evaluated implementations.}
\label{tab:benchmark_scope}
\begin{tabular}{@{}>{\raggedright\arraybackslash}p{0.14\textwidth}
  >{\raggedleft\arraybackslash}p{0.08\textwidth}
  >{\raggedright\arraybackslash}p{0.13\textwidth}
  >{\raggedright\arraybackslash}p{0.32\textwidth}
  >{\raggedright\arraybackslash}p{0.25\textwidth}@{}}
\toprule
\textbf{Workload} & \textbf{Sources} & \textbf{Context} & \textbf{Task and evidence} & \textbf{Evaluation metric} \\
\midrule
RULER-8K & \num{1100} & 8K & Controlled retrieval, tracking, and extraction; generator-derived spans & Exact answer / task score \\
RULER-16K & \num{1100} & 16K & Same controlled families at longer target context; generator-derived spans & Exact answer / task score \\
Qasper & \num{200} & Long document & Document QA; answer-supporting evidence annotations & Normalized reference containment \\
HotpotQA & \num{200} & Multi-document & Multi-hop QA; supporting-fact sentence spans & Normalized reference containment \\
\bottomrule
\end{tabular}
\vspace{2pt}

\begin{tabular}{@{}>{\raggedright\arraybackslash}p{0.29\textwidth}
  >{\raggedright\arraybackslash}p{0.22\textwidth}
  >{\raggedright\arraybackslash}p{0.43\textwidth}@{}}
\toprule
\textbf{Evaluated implementation(s)} & \textbf{Transformed unit} & \textbf{Supported cache measurements} \\
\midrule
StreamingLLM, SnapKV, TOVA, KeyDiff & Token positions or clusters & Measured original-token coverage \\
AdaKV & Head-wise token slots & Measured coverage with head slots kept separate \\
ThinK & Key channels & Structural position addressability; ERR/ECov \texttt{N/A} \\
QuantizedCache (HQQ) & Key/value tensor precision & Structural position addressability; ERR/ECov \texttt{N/A} \\
ChunkKV & Semantic chunks & Projected chunk-to-token coverage \\
\bottomrule
\end{tabular}
\end{table*}
\input{sections/benchmark_contracts}

\subsection{Evaluation on All Sources and Failure Analysis}\label{sec:all_source_evaluation}
We first evaluate every source, then analyze failures without changing the source set.
\input{sections/overview_figure}
Let $\mathcal{X}_d$ be workload $d$'s fixed benchmark source split, $S_i^{F}$ the FullCache score, and $S_{i,m,b}^{C}$ the compressed score for method $m$ at compression setting $b$. FullCache is executed once per source. Every supported $(m,b)$ cell then runs on every $i\in\mathcal{X}_d$ with the same prompt, tokenizer, decoder, and scorer. Retention and channel-pruning methods use the 75/50/25 labels, while QuantizedCache uses 8b/4b/2b; Table~\ref{tab:method_configs} defines the corresponding retained positions, key channels, or bit width, and no byte equivalence is assumed across mechanisms.\looseness=-1

\noindent\textbf{Evaluation on all sources.}
Conventional quality is computed before any diagnostic selection:
\[
Q_{d,m,b}=\frac{1}{|\mathcal{X}_d|}
 \sum_{i\in\mathcal{X}_d}S_{i,m,b}^{C}.
\]
We additionally store the four correctness transitions, C$\rightarrow$C, C$\rightarrow$W, W$\rightarrow$W, and W$\rightarrow$C. The compression-induced failure rate uses all FullCache-correct source instances in the supported cell:
\[
r_{d,m,b}^{CW}=
\frac{|\mathcal{F}_{d,m,b}|}
{|\{i\in\mathcal{X}_d:\mathrm{correct}_i^F=1\}|}.
\]
Across workloads, both values are source weighted. Pooled quality averages all source scores; pooled C$\rightarrow$W rate sums failures and divides by the corresponding FullCache-correct supported pairs.
These results show how often a compressor succeeds and remain comparable with an ordinary benchmark report.\looseness=-1

\noindent\textbf{Paired failure analysis.}
Only after every supported run is complete do we extract
\[
\mathcal{F}_{d,m,b}=\{i\in\mathcal{X}_d:
\mathrm{correct}_i^F=1\land
\mathrm{correct}_{i,m,b}^C=0\}.
\]
Diagnostic means are taken over $\mathcal{F}_{d,m,b}$, not $\mathcal{X}_d$. They profile observed failures, not global quality or method rankings. This order prevents cross-method selection bias.\looseness=-1

\noindent\textbf{Method-specific and matched views.}
A method's own $\mathcal{F}_{d,m,b}$ is the correct view for asking what it breaks. For diagnostic $g$, let $\mathcal A^g_{d,m,b}\subseteq\mathcal X_d$ contain sources with the required paired trace, metadata, defined denominator, and metric-specific integrity check. A direct comparison between $(m,b_m)$ and $(n,b_n)$ uses
$\mathcal I^g=\mathcal F_{d,m,b_m}\cap\mathcal F_{d,n,b_n}\cap\mathcal A^g_{d,m,b_m}\cap\mathcal A^g_{d,n,b_n}$.
The setting pair must have a stated basis for comparison; equal labels alone do not imply equal bytes. Each matched comparison reports intersection, union, and Jaccard. Low Jaccard---not a small intersection alone---shows that the methods fail on different sources.\looseness=-1

\subsection{Run Records and Coverage}\label{sec:kvdench_construction}
Every run record stores identifiers, code and configuration versions, the configured setting and its actual method parameter, and a run status. Supported runs add paired raw and normalized outputs, scorer values, and correctness transitions; C$\rightarrow$W failure rows may also include support metadata and available diagnostics. Retained units map back to original token positions only when the method provides a valid direct or projected mapping. ThinK and QuantizedCache instead record structural position addressability, while ERR and ECov remain \texttt{N/A}; unsupported settings and diagnostics without a valid trace likewise record \texttt{N/A} with a reason.\looseness=-1

Table~\ref{tab:completed_run_accounting} accounts for all \num{2600} sources and \num{62400} records. Excluding one unsupported ThinK/25\% cell per source leaves \num{59800} completed runs; FullCache correctness defines C$\rightarrow$W eligibility.\looseness=-1
\begin{table}[t]
\centering
\scriptsize
\setlength{\tabcolsep}{2.2pt}
\renewcommand{\arraystretch}{0.96}
\caption{Run accounting. \texttt{N/A} denotes ThinK/25\%; FC-eligible and C$\rightarrow$W count FullCache-correct supported pairs, and Unique counts affected sources.}
\label{tab:completed_run_accounting}
\begin{tabular*}{\columnwidth}{@{\extracolsep{\fill}}lrrrrrrr@{}}
\toprule
\textbf{Workload} & \textbf{Src.} & \textbf{Records} & \textbf{N/A} & \textbf{Runs} & \textbf{FC-elig.} & \textbf{C$\rightarrow$W} & \textbf{Unique} \\
\midrule
RULER-8K  & \num{1100} & \num{26400} & \num{1100} & \num{25300} & \num{22264} & \num{5970} & \num{968} \\
RULER-16K & \num{1100} & \num{26400} & \num{1100} & \num{25300} & \num{21873} & \num{5396} & \num{951} \\
Qasper    & \num{200}  & \num{4800}  & \num{200}  & \num{4600}  & \num{1357}  & \num{327}  & \num{49} \\
HotpotQA  & \num{200}  & \num{4800}  & \num{200}  & \num{4600}  & \num{3404}  & \num{827}  & \num{126} \\
\midrule
\textbf{Total} & \textbf{\num{2600}} & \textbf{\num{62400}} & \textbf{\num{2600}} & \textbf{\num{59800}} & \textbf{\num{48898}} & \textbf{\num{12520}} & \textbf{\num{2094}} \\
\bottomrule
\end{tabular*}
\end{table}

\input{sections/benchmark_use}

The record format also accommodates new compressors through supported diagnostics and new tasks through a scorer and optional support evidence, without changing existing rows, setting semantics, or failure selection.\looseness=-1

%% file: sections/benchmark_contracts.tex
\subsection{Task and Method Adapters}\label{sec:task_method_adapters}
A task adapter applies documented source-inclusion rules, then fixes the source identifier, prompt template, model input, reference answer, scorer version, and context length before compression. It aligns evidence after final-prompt tokenization because source offsets can shift when instructions and chat formatting are added. RULER generators provide exact answer-bearing spans; Qasper and HotpotQA map official support text into the prompt and record alignment success. Once a split is fixed, missing or ambiguous mappings never remove a source; they only make the affected cache diagnostic unavailable during paired failure analysis.\looseness=-1

A method adapter receives the same tokenized prompt and returns status, configured setting, actual method parameter, output, scorer input, and code/configuration versions. When applicable, it also maps retained token, block, or chunk identifiers to original positions. Likelihood traces use the same reference tokens under FullCache and compression. Attention traces use a separate eager-mode run only when compression behavior is unchanged, so inspectability does not confer an evaluation advantage.

Record keys contain workload, source, model revision, method, setting, and run configuration. Aggregation verifies unique keys, FullCache partners, prompt hashes, numeric fields, and complete fixed-split coverage. Execution errors are audited separately rather than scored as wrong answers, and unsupported settings remain distinct from execution errors.\looseness=-1

%% file: sections/overview_figure.tex
\begin{figure}[!t]
\centering
\begin{tikzpicture}[
    x=1cm,
    y=1cm,
    font=\fontsize{8.4pt}{9.2pt}\selectfont,
    flow/.style={-{Latex[length=1.90mm,width=1.30mm]},
        draw=blue!52!black, line width=0.72pt,
        line cap=round, line join=round},
    stage title/.style={font=\fontsize{9.1pt}{9.8pt}\selectfont\bfseries,
        text=black!88, align=center},
    note/.style={font=\fontsize{7.8pt}{8.5pt}\selectfont,
        text=black!72, align=center, text width=7.55cm},
    pair/.style={draw=black!55, rounded corners=2pt, line width=0.52pt,
        align=center, text width=3.00cm, minimum height=0.82cm,
        inner xsep=3pt, inner ysep=1.5pt},
    band/.style={draw=black!55, rounded corners=2pt, line width=0.55pt,
        align=center, text width=7.52cm, minimum height=0.84cm,
        inner xsep=3pt, inner ysep=1.5pt},
    gate/.style={draw=orange!75!black, fill=orange!10, rounded corners=2pt,
        line width=0.52pt, align=center, text width=6.55cm,
        minimum height=0.58cm, inner xsep=3pt, inner ysep=1.5pt},
    probe/.style={draw=black!50, rounded corners=2pt, line width=0.48pt,
        align=center, text width=1.46cm, minimum height=0.57cm,
        inner xsep=2.5pt, inner ysep=1pt},
    output/.style={draw=black!60, rounded corners=2pt, line width=0.55pt,
        fill=black!3, align=center, text width=3.27cm,
        minimum height=0.68cm, inner xsep=3pt, inner ysep=1pt},
    group frame/.style={draw=black!35, rounded corners=2pt,
        line width=0.45pt}
]

\node[stage title] (inputtitle)
  {1\quad Fixed run setup and verified adapters};

\node[pair, fill=black!3, below=0.06cm of inputtitle, xshift=-1.84cm] (full)
  {\textbf{FullCache control}\\once per source\\reused across cells};
\node[pair, fill=blue!6, below=0.06cm of inputtitle, xshift=1.84cm] (compressed)
  {\textbf{Compressed matrix}\\supported $(m,b)$ cells\\status recorded per run};

\node[draw=none, fit=(full)(compressed), inner sep=0pt] (inputrow) {};

\node[band, fill=teal!8, below=0.20cm of inputrow] (fullmatrix)
  {\textbf{2\quad Results for all runs}\\task score $\mid$ four transitions $\mid$ C$\rightarrow$W rate\\denominator: all FullCache-correct sources in the cell};
\draw[flow] (full.south) -- ([xshift=-0.95cm]fullmatrix.north);
\draw[flow] (compressed.south) -- ([xshift=0.95cm]fullmatrix.north);

\node[gate, below=0.20cm of fullmatrix] (failuregate)
  {\textbf{3\quad Complete all runs, then extract $\mathcal{F}_{d,m,b}$}\\that cell's own C$\rightarrow$W rows};
\draw[flow] (fullmatrix.south) -- (failuregate.north);

\node[stage title, below=0.20cm of failuregate] (diagtitle)
  {4\quad Available paired diagnostics};
\node[probe, fill=blue!7, below=0.02cm of diagtitle, xshift=-2.82cm] (cache)
  {\textbf{Cache}\\mapped retention};
\node[probe, fill=violet!8, below=0.02cm of diagtitle, xshift=-0.94cm] (likelihood)
  {\textbf{Likelihood}\\predictive drift};
\node[probe, fill=green!8, below=0.02cm of diagtitle, xshift=0.94cm] (attn)
  {\textbf{Attention}\\support access};
\node[probe, fill=red!6, below=0.02cm of diagtitle, xshift=2.82cm] (decode)
  {\textbf{Decoding}\\output / scorer};
\node[group frame, fit=(diagtitle)(cache)(decode),
      inner xsep=4pt, inner ysep=2.5pt] (diagnostics) {};
\draw[flow] (failuregate.south) -- (diagnostics.north);

\node[stage title, below=0.18cm of diagnostics] (reporttitle)
  {5\quad Reporting views};
\node[output, below=0.02cm of reporttitle, xshift=-1.82cm] (own)
  {\textbf{Method failure profile}\\coverage of $\mathcal{F}_{d,m,b}$};
\node[output, below=0.02cm of reporttitle, xshift=1.82cm] (matched)
  {\textbf{Controlled comparison}\\intersection + sizes + Jaccard};
\node[group frame, fit=(reporttitle)(own)(matched),
      inner xsep=4pt, inner ysep=2.5pt] (reporting) {};
\draw[flow] (diagnostics.south) -- (reporting.north);

\end{tikzpicture}
\caption{All supported runs precede cell-wise C$\rightarrow$W selection and diagnostics. Method-specific sets preserve coverage; intersections support controlled comparisons.}
\Description{Fixed inputs feed one reusable FullCache control and all supported compressed method-setting cells into a complete run matrix. Only after all runs are complete are each cell's compression-induced failures extracted, linked to available cache, likelihood, attention, and decoding diagnostics, and reported through method-specific or fixed-intersection views.}
\label{fig:overview}
\end{figure}
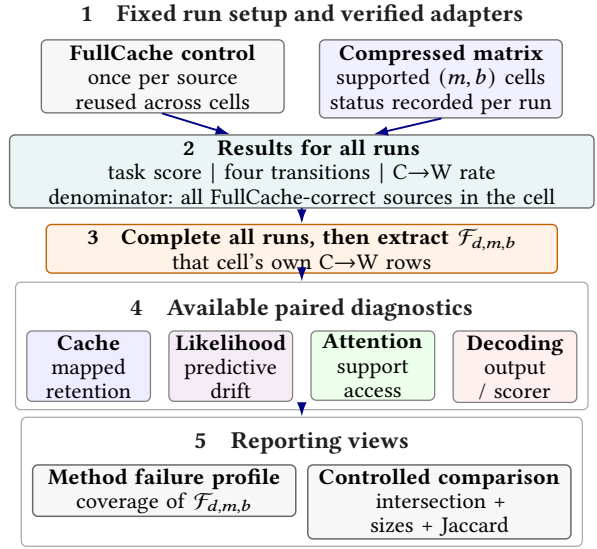

%% file: sections/benchmark_use.tex
% !TEX root = ../main.tex
\subsection{Benchmark Use and Reporting}\label{sec:benchmark_use}
\textsc{KVDiagnosis} treats signatures as component-specific routes, not causal labels; the low-EAR intervention, sham, and C$\rightarrow$C safety control validate one route. Reports pair task quality and C$\rightarrow$W rate with cache representation, supported and \texttt{N/A} counts, valid denominators, and method-specific failure sets. Settings rank compression only within a method, and matched comparisons report overlap; distinct failures are not collapsed into a composite score.\looseness=-1

Appendix Table~\ref{tab:artifact_components} specifies the versioned release files and consistency checks; each published C$\rightarrow$W row links to its FullCache pair, diagnostics, and generator.

%% file: sections/metirc.tex
% !TEX root = ../main.tex
\section{Diagnostic Metrics}\label{sec:metric}

A C$\rightarrow$W transition establishes that compression changed a correct answer; it does not locate the change. Prior sample- and instruction-level studies expose this gap but do not provide a shared, structured failure record~\cite{rethinking_kv_cache_compression,chen2026pitfalls}. We therefore ask whether the evidence was retained, whether its effect on next-token predictions was preserved, whether generation could access it, and whether the output changed only during decoding or scoring. These measurements yield component-specific signatures; Section~\ref{sec:analysis} evaluates their failure--success separation and tests the low-EAR access signature with a paired intervention.\looseness=-1

\noindent\textbf{Cache retention.}
Let $z=(d,i,m,b)$ index workload $d$, source $i$, method $m$, and setting $b$. Let $E_{i,1},\ldots,E_{i,J_i}$ be the nonempty evidence spans and $E_i=\bigcup_jE_{i,j}$ their token positions. When a valid original-token map exists, $R_{z,u}$ is the set represented after compression in layer--KV-head slot $u\in\mathcal U_z$. Let $c_{z,u,j}=|R_{z,u}\cap E_{i,j}|/|E_{i,j}|$ and set $\tau=0.5$.
\begingroup
\setlength{\abovedisplayskip}{1pt}
\setlength{\belowdisplayskip}{1pt}
\setlength{\abovedisplayshortskip}{1pt}
\setlength{\belowdisplayshortskip}{1pt}
\small
\[
\begin{aligned}
\mathrm{ERR}_{z}&=\frac{1}{|\mathcal U_z|}\sum_{u\in\mathcal U_z}
 \frac{|R_{z,u}\cap E_i|}{|E_i|},\\[-2pt]
\mathrm{ECov}_{z}&=\frac{1}{|\mathcal U_z|J_i}\sum_{u\in\mathcal U_z}\sum_{j=1}^{J_i}
 \mathbf{1}[c_{z,u,j}\ge\tau].
\end{aligned}
\]
\endgroup
Slot averaging prevents a cross-head union from producing artificial perfect coverage. We record four applicability states: \emph{measured token coverage} for direct original-position maps, \emph{projected token coverage} for ChunkKV's chunk-to-token map, \emph{structural position addressability} for ThinK and QuantizedCache, and \texttt{N/A} when none applies. ERR and ECov are numeric only in the first two states. Structural addressability is recorded separately; annotated evidence positions remain addressable, but representation fidelity is unknown.\looseness=-1

\noindent\textbf{Predictive distribution.}
For reference tokens $y_{i,1:T_i}$, let $p^F_{d,i,t}$ and $p^C_{z,t}$ be the paired FullCache and compressed teacher-forced next-token distributions, and let $\epsilon>0$ be a small constant for numerical stability. FullCache is independent of $m,b$, whereas its compressed partner is not:
\begingroup
\small
\[
\begin{aligned}
\mathrm{NLL}^{F}_{d,i}&=-\frac{1}{T_i}\sum_t\log p^F_{d,i,t}(y_{i,t}),\\
\mathrm{NLL}^{C}_{z}&=-\frac{1}{T_i}\sum_t\log p^C_{z,t}(y_{i,t}),\\
\Delta\mathrm{NLL}_{z}&=\mathrm{NLL}^{C}_{z}-\mathrm{NLL}^{F}_{d,i},\qquad
\mathrm{GPR}_{z}=\frac{e^{-\mathrm{NLL}^{C}_{z}}}{e^{-\mathrm{NLL}^{F}_{d,i}}+\epsilon}.
\end{aligned}
\]
\endgroup
Full-to-compressed KL, Top-50 overlap, and gold-token rank shift provide complementary distribution checks; Figure~\ref{fig:c2c_separation} uses one minus Top-50 overlap as the corresponding failure-risk direction. Positive $\Delta$NLL, low GPR/Top-50, or positive rank shift records paired drift, not its cause.\looseness=-1

\noindent\textbf{Access, decoding, and valid rows.}
On a valid eager trace that maps back to original positions, evidence attention mass (EAM) averages answer-time attention assigned to evidence over the common layer--head--answer-step domain. Evidence attention retention (EAR) divides compressed EAM by FullCache EAM on that same domain, and normalized evidence attention enrichment (NEAE) divides EAM by the evidence share of the original context. Decoding records first/mean gold rank, evaluated and emitted tokens, stop reason, and normalized output. For every diagnostic $q$, let $\mathcal A^q_{d,m,b}\subseteq\mathcal X_d$ contain the sources with the required paired trace, metadata, defined denominator, and metric-specific validity check. The mean uses only failure rows on which $q$ is valid:
\begingroup
\small
\[
\overline q_{d,m,b}=
\frac{\sum_{i\in\mathcal F_{d,m,b}\cap\mathcal A^q_{d,m,b}}q_{d,i,m,b}}
{|\mathcal F_{d,m,b}\cap\mathcal A^q_{d,m,b}|}.
\]
\endgroup
Missing values are not imputed or treated as zero; an empty intersection is \texttt{N/A}. The definitions above specify each reported quantity and its validity conditions; Appendix~\ref{app:ruler8k_complete} applies them to complete method--setting results.\looseness=-1

%% file: sections/experiment.tex
% !TEX root = ../main.tex
\section{Experimental Evaluation}\label{sec:analysis}

We ask four questions: failure frequency (RQ1), diagnostic separation and targeted repair (RQ2), differences hidden by similar aggregate scores (RQ3), and transfer across models on RULER and evidence-annotated QA (RQ4). Scores and C$\rightarrow$W denominators precede failure analysis.

\subsection{Experimental Setup}
\noindent\textbf{Implementation details.}
Qwen3-8B is the primary model; cross-model analysis uses Falcon3-7B-Instruct and Mistral-Small-24B-Instruct-2501 on RULER-16K, Qasper, and HotpotQA. Paired runs share the prompt, tokenizer, reference, scorer, and greedy decoder; only the cache path changes. FullCache runs once per source as the paired control, not a ninth compressor, and cells match source IDs and prompt hashes. Runs use one NVIDIA H200 in bf16 with PyTorch 2.11.0+cu128, Transformers 5.2.0, and \texttt{kvpress} 0.5.3. Appendix Table~\ref{tab:environment} records exact model revisions and software. Each fixed cell has one deterministic result, so figures omit error bars.

\noindent\textbf{Datasets and metrics.}
RULER-8K and RULER-16K each contain \num{1100} instances spanning needle retrieval, multi-key and multi-value retrieval, variable tracking, and word extraction. Their generators expose exact evidence positions and task-specific exact-answer scoring. Qasper and HotpotQA each contribute 200 deterministically sampled, evidence-mapped sources. The QA scorer is benchmark-specific rather than official: it normalizes output and references, awards the fraction of references found as substrings, and marks correctness only at 1.0; multiple-choice items use exact option matching. Appendix~\ref{app:completed_extra_results} gives selection and mapping details.

\noindent\textbf{Methods and settings.}
We test StreamingLLM, SnapKV, TOVA, KeyDiff, ThinK, ChunkKV, AdaKV, and QuantizedCache; Figure~\ref{fig:taxonomy} and Table~\ref{tab:benchmark_scope} map their mechanism families and cache representations. Eviction, chunk, allocation, and ThinK methods use 75\%, 50\%, and 25\% settings; QuantizedCache uses HQQ 8-, 4-, and 2-bit tensors (50\%, 25\%, and 12.5\% ideal bit retention versus bf16). Settings therefore order compression only within a method. ThinK/25\% is \texttt{N/A}; every other cell covers the fixed split. Latency, throughput, bytes, and energy are outside scope.\looseness=-1

\begin{figure}[t]
\centering
\includegraphics[width=\columnwidth]{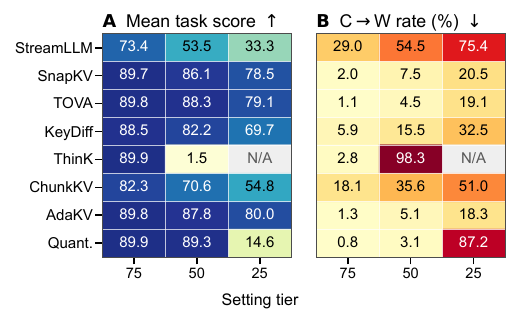}
\vspace{-7pt}
\caption{All \num{2600} sources: (A) mean task score; (B) C$\rightarrow$W rate. Gray marks unsupported ThinK/25\%; QuantizedCache uses 8b/4b/2b, while other methods use 75/50/25.}
\Description{Two heatmaps show mean task score and compression-induced failure rate for eight methods over all sources.}
\label{fig:population_outcomes}
\end{figure}

\begin{figure}[t]
\centering
\includegraphics[width=\columnwidth]{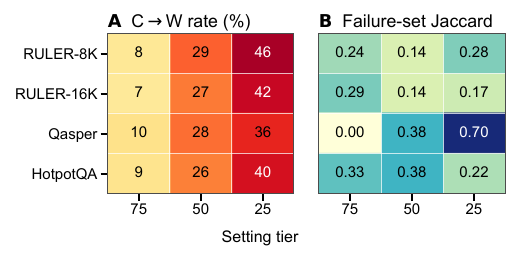}
\vspace{-7pt}
\caption{Failure frequency and identity: (A) workload-level C$\rightarrow$W rate; (B) SnapKV--TOVA failure-set Jaccard. Columns are light/intermediate/aggressive settings: 75/50/25\% for retention methods and 8b/4b/2b for QuantizedCache in panel A.}
\Description{Two heatmaps show workload-level compression-induced failure rates and SnapKV--TOVA failure-set overlap across three settings.}
\label{fig:population_overlap}
\end{figure}

\subsection{RQ1: How Often Are Correct Answers Lost?}
Figure~\ref{fig:population_outcomes} reports mean score over \num{2600} sources (A) and C$\rightarrow$W rate among FullCache-correct sources (B). FullCache scores 90.0 with 81.8\% binary accuracy. At the light tier, TOVA, AdaKV, SnapKV, and QuantizedCache stay within 0.3 points, but their C$\rightarrow$W rates span 0.8--2.0\%---a 2.5$\times$ difference hidden by aggregate quality. At the aggressive tier, TOVA/AdaKV score 79.1/80.0 versus 33.3/54.8 for StreamingLLM/ChunkKV. QuantizedCache falls from 89.3 at 4b to 14.6 at 2b; ThinK collapses at 50\% and has no 25\% result.

Figure~\ref{fig:population_overlap} resolves failure frequency and identity by workload. Panel A shows the C$\rightarrow$W rate rising from 8/7/10/9\% to 46/42/36/40\% on RULER-8K/RULER-16K/Qasper/HotpotQA as the setting moves from the light to the aggressive tier (75\% to 25\%, or 8b to 2b for QuantizedCache). Panel B shows SnapKV--TOVA Jaccard overlap at or below 0.38 in 11 of 12 workload--setting cells; the exception is Qasper/25\% at 0.70. Thus aggregate quality, paired failure frequency, and failure identity answer distinct questions.\looseness=-1

\begin{figure}[t]
\centering
\includegraphics[width=.75\columnwidth]{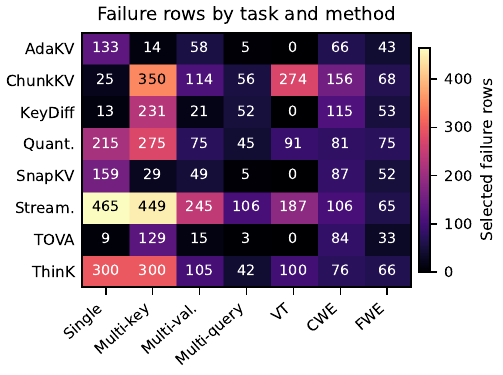}
\vspace{-7pt}
\caption{RULER-8K failures by task and method. Counts reveal method-specific task concentration but do not rank task difficulty.}
\Description{A heatmap shows RULER task families by eight compression methods, with failure-row counts in each cell.}
\label{fig:ruler8k_task_method_matrix}
\end{figure}

\noindent\textbf{Failure composition is method-specific.}
Figure~\ref{fig:ruler8k_task_method_matrix} shows distinct task profiles: StreamingLLM concentrates on single- and multi-key retrieval, ChunkKV on multi-key retrieval and variable tracking, whereas QuantizedCache is more diffuse. Because task-family sizes differ, these are composition counts, not difficulty estimates.

\noindent\textbf{Finding 1.} Pooled failure frequency rises from the light to the aggressive setting tier on every workload, but method-level trends vary and ThinK is absent at 25\%; this is not a cross-mechanism ranking.

\begin{table*}[t]
\centering
\fontsize{8.5pt}{9.5pt}\selectfont
\setlength{\tabcolsep}{3.5pt}
\caption{Predefined diagnostic categories for 12,520 C$\rightarrow$W method--setting rows. A match to exactly one of the first six rules assigns its named category; multiple matches assign \emph{Conflicting}, and no match assigns \emph{Ambiguous}. Sources may repeat. EAR/Top-50 are defined for 7,038 (56.2\%)/8,400 (67.1\%) rows; missing traces are not imputed.}
\label{tab:failure_regimes}
\begin{tabular}{@{}p{0.215\textwidth}p{0.305\textwidth}rrp{0.30\textwidth}@{}}
\toprule
\textbf{Diagnostic category} & \textbf{Rule} & \textbf{Rows} & \textbf{Share} & \textbf{Diagnostic interpretation} \\
\midrule
Low mapped coverage & mapped ECov$<0.50$ & 5,047 & 40.3\% & Broad measured/projected support loss \\
Partial mapped coverage & $0.50\leq$ mapped ECov$<0.90$ & 2,866 & 22.9\% & Incomplete measured/projected support \\
High mapped-coverage drift & mapped ECov$\geq0.90$, $\Delta$NLL$\geq1$ & 19 & 0.2\% & Drift despite mapped position retention \\
Structural-position drift & structural addressability, $\Delta$NLL$\geq1$ & 2,126 & 17.0\% & Positions addressable; fidelity unknown \\
Low-EAR candidate & mapped ECov$\geq0.90$ or structural addressability; EAR$<0.50$ & 104 & 0.8\% & Evidence-access or routing deficit \\
Decoding/scoring candidate & mapped ECov$\geq0.90$ or structural addressability; $|\Delta$NLL$|\leq0.10$, Top-50$\geq0.90$ if available & 405 & 3.2\% & Downstream output/scorer sensitivity \\
Conflicting diagnostic signals & Two or more rules match & 1,556 & 12.4\% & Multiple signals co-occur \\
Ambiguous & No rule matches & 397 & 3.2\% & No predefined signature \\
\bottomrule
\end{tabular}
\end{table*}

\subsection{RQ2: What Characterizes Failures?}
The resource contains \num{12520} C$\rightarrow$W method--setting rows; pooled shares count rows because sources may recur. Six predeclared rules yield mutually exclusive categories, with multiple or no matches labeled \emph{Conflicting} or \emph{Ambiguous}. Coverage rules require measured token maps; structural addressability remains separate, and missing fields cannot trigger a rule. The categories are observational signatures; intervention evidence is reported separately.

\begin{figure}[t]
\centering
\includegraphics[width=.97\columnwidth]{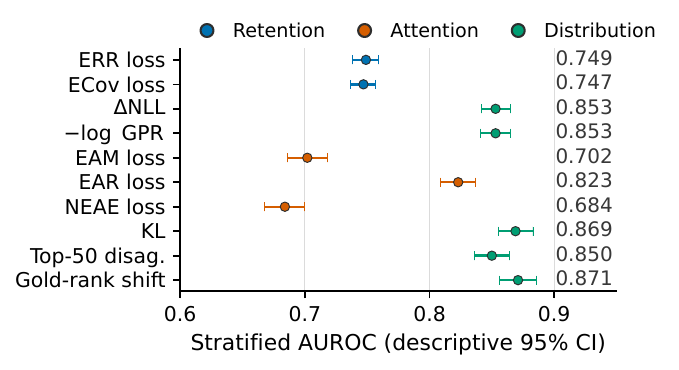}
\caption{C$\rightarrow$C versus C$\rightarrow$W diagnostic separation. Larger values indicate greater failure risk; ERR/ECov exclude structural methods. Metrics use only valid mixed strata, and intervals are descriptive because sources recur.}
\Description{A forest plot reports stratified AUROC and descriptive 95 percent intervals for ten cache-retention, likelihood, attention, and rank diagnostics. Every estimate is above the random baseline of 0.5.}
\label{fig:c2c_separation}
\end{figure}

\noindent\textbf{Diagnostics separate failures from successful compression.}
Figure~\ref{fig:c2c_separation} compares C$\rightarrow$W with a ledger of 4,936 C$\rightarrow$C runs, using only mixed strata and valid measurements. All ten predefined failure-risk directions exceed random ranking: stratified AUROC ranges from 0.684 for NEAE loss to 0.871 for gold-rank shift. KL (0.869), $\Delta$NLL (0.853), Top-50 disagreement (0.850), and EAR loss (0.823) are strongest. After excluding structural rows, ERR loss (0.749) and ECov loss (0.747) also separate successful from failed compression. EAM and NEAE are weaker and vary more by method. Thus the diagnostics are not properties shared by every compressed run.

\begin{figure}[!t]
\centering
\includegraphics[width=\columnwidth]{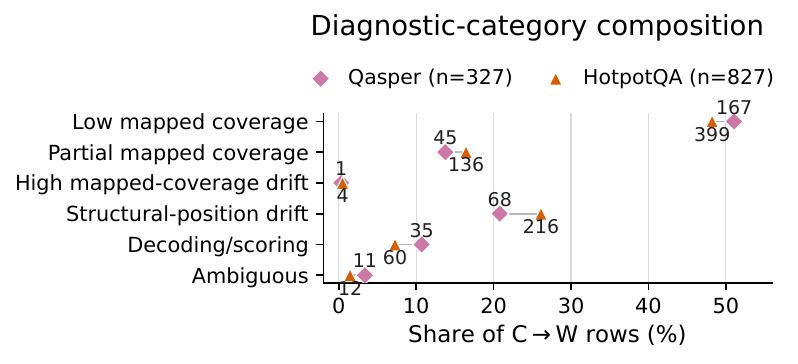}
\caption{Diagnostic-category composition on evidence-annotated QA. Points report within-workload shares; labels give C$\rightarrow$W row counts. Neither subset has conflicting labels or valid eager-attention traces.}
\Description{A paired-dot plot compares six diagnostic-category shares and exact failure-row counts for Qasper and HotpotQA.}
\label{fig:qa_transfer}
\end{figure}

\noindent\textbf{Evidence coverage is often low or partial.}
The layer--KV-head estimator in Section~\ref{sec:metric} avoids a saturating cross-slot union. Among all failure rows, 5,047 (40.3\%) have low mapped coverage and 2,866 (22.9\%) have partial mapped coverage. Figure~\ref{fig:qa_transfer} shows the same combined share on evidence-annotated QA: 212/327 Qasper failures (64.8\%) and 535/827 HotpotQA failures (64.7\%). Lower retention also reduces ECov within position-selection methods: StreamingLLM falls from 0.262 to 0.074 and TOVA from 0.951 to 0.445 between 75\% and 25\%. These values describe a typical layer--head slot, not global absence of every support-span copy.

\begin{figure}[!t]
\centering
\includegraphics[width=\columnwidth]{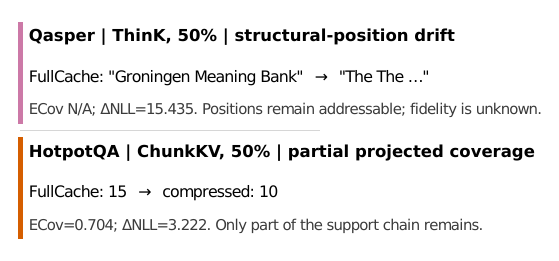}
\caption{Paired QA cases. ThinK preserves structural addressability but incurs severe likelihood drift (top); ChunkKV retains partial projected evidence coverage and changes a numerical answer (bottom).}
\Description{Two paired cases contrast structural likelihood drift on Qasper with partial evidence coverage on HotpotQA.}
\label{fig:qa_cases}
\end{figure}

\begin{figure*}[!t]
\centering
\begin{tikzpicture}
\node[inner sep=0,anchor=south west] (crossmodel) at (0,0)
  {\includegraphics[width=.92\textwidth]{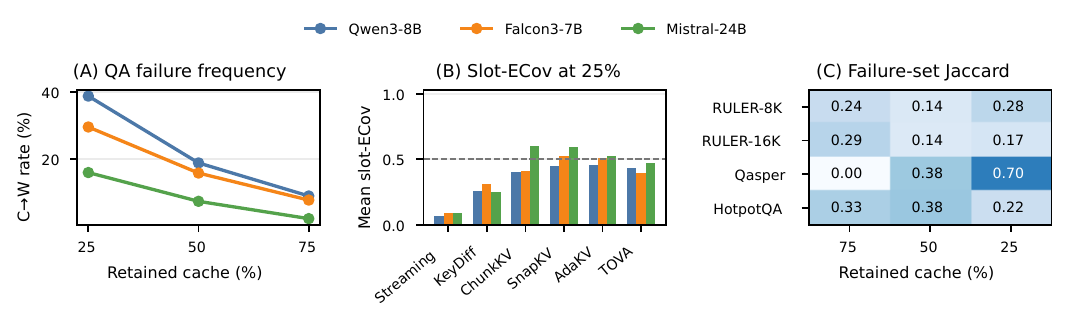}};
\begin{scope}[x={(crossmodel.south east)},y={(crossmodel.north west)}]
  \fill[white] (0.075,0.075) rectangle (0.295,0.195);
  \node[font=\scriptsize] at (0.185,0.135) {Setting tier};
\end{scope}
\end{tikzpicture}
\vspace{-5pt}
\caption{Cross-model validation. (A) QA C$\rightarrow$W rate pooled over six position methods and QuantizedCache; the setting tiers map 75/50/25\% to 8b/4b/2b for QuantizedCache. (B) Mean slot-ECov over RULER-16K/QA position-method failures at 25\% retention; dashed: low-coverage threshold. (C) Qwen SnapKV--TOVA failure-set Jaccard. Comparisons use model-specific denominators and slot geometry.}
\Description{Panels A--B compare Qwen, Falcon, and Mistral failure frequency and coverage. Panel C shows that failure identities differ between methods and settings.}
\label{fig:crossmodel_validation}
\end{figure*}

\noindent\textbf{Structural addressability is not measured coverage.}
Only 19 rows (0.2\%) combine mapped ECov of at least 0.90 with a gold-answer NLL increase of at least one. Separately, 2,126 rows (17.0\%) show the same likelihood drift under ThinK or QuantizedCache. For these methods, annotated evidence positions remain addressable, but representation fidelity is unknown; ERR and ECov are therefore \texttt{N/A}. On RULER-16K, ThinK at 50\% flips all 951 FullCache-correct sources with mean $\Delta$NLL 7.958 and GPR 0.015. QuantizedCache at 2b flips 883 of 951 with $\Delta$NLL 2.230 and GPR 0.260. In the Qasper case (Figure~\ref{fig:qa_cases}), ThinK changes "Groningen Meaning Bank" into a repeated "The" sequence while $\Delta$NLL reaches 15.435. Position addressability cannot replace a fidelity or likelihood check.

\noindent\textbf{Low EAR predicts selective repair.}
The released records provide valid paired eager-attention traces and EAR for 7,038 rows (56.2\%). Of the 104 rows matching only the low-EAR rule, 96 reproduce the failure under the intervention's eager-attention baseline. Holding the compressed cache and decoder fixed, adding $\log 4$ to the attention logits of retained gold-evidence positions repairs 28/96 (29.2\%); boosting an equal number of deterministic non-evidence positions repairs 6/96 (6.3\%), a paired difference of 22.9 percentage points (23 evidence-only versus 1 sham-only repair; 95\% paired-bootstrap CI, 14.6--32.3 points; two-sided exact McNemar $p=2.98\!\times\!10^{-6}$). The same evidence boost degrades only 3/92 (3.3\%) matched C$\rightarrow$C controls. This selective response supports low EAR as an access/routing signature in this cohort. Separately, Top-50 is valid for 8,400 rows (67.1\%); 405 position-addressable, stable-likelihood rows remain decoding/scoring candidates, while 1,556 rows have conflicting signals and 397 are ambiguous.

\noindent\textbf{Finding 2.} Matched controls show that all ten diagnostics separate C$\rightarrow$W from C$\rightarrow$C; low EAR also predicts selective evidence-access repair. Low or partial mapped coverage is most frequent. Structural addressability must be paired with fidelity diagnostics rather than interpreted as coverage.\looseness=-1

\subsection{RQ3: What Do Similar Scores Hide?}
SnapKV and TOVA each contribute \num{7800} compressed runs (\num{2600} sources at three settings), with mean scores of 84.8 and 85.7. Yet 11 of 12 workload--setting cells have failure-set Jaccard overlap at or below 0.385 (Figure~\ref{fig:population_overlap}B): similar aggregate quality does not imply similar failures. Qasper rises to 0.70 at 25\%, whereas HotpotQA falls from 0.38 to 0.22. Method-specific failure sets are therefore necessary, and intersections must be recomputed for each setting.

\noindent\textbf{Finding 3.} Similar aggregate quality does not imply interchangeable failures. Full method-specific sets preserve coverage; intersections support controlled comparisons.\looseness=-1

\subsection{RQ4: Do Failure Signatures Persist Across Models?}
We repeat paired C$\rightarrow$W selection on Falcon3-7B and Mistral-24B with model-native geometry. Across Qwen, Falcon, and Mistral, the coverage analysis contains \num{13597} mapped RULER-16K/QA position-method failures (Figure~\ref{fig:crossmodel_validation}).

\noindent\textbf{Coverage diagnoses transfer.}
From the light to the aggressive tier (75\% to 25\%, or 8b to 2b for QuantizedCache), the pooled QA C$\rightarrow$W rate in Figure~\ref{fig:crossmodel_validation}A rises from 8.9\% to 38.7\% (Qwen), 7.7\% to 29.5\% (Falcon), and 2.1\% to 15.9\% (Mistral). Across the 18 position-method setting cells in panel B, slot-ECov correlates with Qwen at $\rho=0.969$ for Falcon and $0.957$ for Mistral (both $p<10^{-9}$), and all six methods decline monotonically. StreamingLLM shows the clearest 25\% coverage loss (0.069/0.087/0.091); SnapKV and AdaKV retain more support.

\noindent\textbf{Fidelity drift is model-dependent.}
On RULER-16K, failure-row mean $\Delta$NLL pooled over each method's supported settings is 2.113, 2.177, and 1.760 for StreamingLLM and 7.819, 4.963, and 2.950 for ThinK across Qwen, Falcon, and Mistral. QuantizedCache marks a boundary: 2.153 on Qwen versus 0.348 and 0.021 on Falcon and Mistral. Falcon's ThinK QA configuration lacks a comparable sweep: 50\% collapses and 25\% is unsupported.

\noindent\textbf{Finding 4.} Failure frequency and coverage diagnoses generalize across architectures, whereas quantization remains model-specific. Thus the procedure, not metric magnitude, generalizes.\looseness=-1

\subsection{Validation and Study Limitations}
The released ledger regenerates all \num{12520} C$\rightarrow$W diagnostic keys. Token maps are measured, chunk maps projected, and structural methods or missing traces remain \texttt{N/A}; recurring sources make AUROC intervals descriptive. Cross-model claims cover three logged revisions, greedy decoding, method-specific settings, fixed splits, and trace availability; attention AUROC and interventions remain Qwen-only. The Appendix audits environment, licenses, and AI use.\looseness=-1

%% file: sections/conclusion.tex
% !TEX root = ../main.tex
\section{Conclusion}\label{sec:conclusion}
KV-cache compression enables long-context inference, but aggregate scores cannot explain changed answers. \textsc{KVDiagnosis} pairs every compressed run with FullCache, selects C$\rightarrow$W rows, and reports only valid diagnostics. Across \num{59800} primary-model runs, 63.2\% of \num{12520} failures show low or partial measured/projected coverage; all ten diagnostics separate C$\rightarrow$W from C$\rightarrow$C, and targeted evidence attention repairs 29.2\% of reproducible low-EAR failures versus 6.3\% under a matched sham. Coverage trends reproduce on Falcon3-7B and Mistral-24B, whereas quantization drift remains model-specific. Explicit denominators, native geometry, and missing-value semantics support explaining failures rather than merely ranking compressors.\looseness=-1

%% file: sections/appendix.tex
% !TEX root = ../main.tex

\appendix
\onecolumn
\raggedbottom

\setcounter{topnumber}{4}
\setcounter{dbltopnumber}{4}
\renewcommand{\topfraction}{0.95}
\renewcommand{\dbltopfraction}{0.95}
\renewcommand{\textfraction}{0.03}
\renewcommand{\floatpagefraction}{0.80}
\renewcommand{\dblfloatpagefraction}{0.80}

\section{Reproducibility and Detailed Results}
\label{app:reproducibility_scope}

This appendix records the configurations, data sources, complete per-workload results, and validation checks needed to interpret the main paper. The public release preserves these records in a versioned archive with checksums, recorded dependency versions, runnable commands, validation logs, and compatible licenses.

\paragraph{PyramidKV adapter audit.}
The excluded implementation used \texttt{kvpress} 0.5.3 (source commit \texttt{fa7a0dc}). The two classes were \texttt{SnapKVPress} and \texttt{PyramidKVPress}, from the \texttt{snapkv\_press} and \texttt{pyramidkv\_press} modules. Both used a window of 64 and kernel size 5; PyramidKV additionally used \texttt{beta=20}. The tracker replaced both \texttt{compress} methods with a generic fixed-budget path, bypassing PyramidKV's \texttt{get\_layer\_budget}. Across 3,300 RULER-8K, 3,300 RULER-16K, and 1,200 QA pairs, persisted retained-index mappings, raw and extracted outputs, scores, and gold NLL are equal in all 7,800 pairs. Every pair came from a different Slurm job and 84.2\% from different nodes, excluding a file-copy explanation. Table~\ref{tab:pyramid_adapter_audit} gives representative SHA-256 prefixes; the public audit records full hashes, class signatures, source hashes, configurations, and all equality rates. The rows are excluded rather than relabeled.

\begin{table}[h]
\centering
\small
\setlength{\tabcolsep}{5pt}
\caption{Representative SnapKV/PyramidKV pair audit. Each hash is identical across the two method labels; KV is the retained ratio.}
\label{tab:pyramid_adapter_audit}
\begin{tabular}{@{}llcrrc@{}}
\toprule
\textbf{Workload} & \textbf{Sample} & \textbf{KV} & \textbf{Index hash} & \textbf{Output hash} & \textbf{Equal} \\
\midrule
RULER-8K & \texttt{cwe\_000000} & 75\% & \texttt{514e8b39ca} & \texttt{d1626650b9} & yes \\
RULER-16K & \texttt{cwe\_000000} & 50\% & \texttt{07e043e559} & \texttt{900b8707ae} & yes \\
QA bridge & \texttt{hotpotqa\_000000} & 25\% & \texttt{2b930b7d86} & \texttt{acc60cc7bd} & yes \\
\bottomrule
\end{tabular}
\end{table}

\begingroup
\small
\setlength{\tabcolsep}{2.5pt}
\renewcommand{\arraystretch}{1.04}
\setlength{\LTcapwidth}{\textwidth}
\begin{longtable}{@{}
  >{\raggedright\arraybackslash}p{0.155\textwidth}
  >{\raggedright\arraybackslash}p{0.225\textwidth}
  >{\raggedright\arraybackslash}p{0.145\textwidth}
  >{\raggedright\arraybackslash}p{0.30\textwidth}
  >{\centering\arraybackslash}p{0.10\textwidth}@{}}
\caption{Twenty-five-method survey taxonomy and its connection to the \textsc{KVDiagnosis} diagnostics. \textbf{Evaluated} marks the eight experimental implementations; other rows provide literature coverage.}
\label{tab:method_landscape}\\
\toprule
\textbf{Method} & \textbf{Compression decision} & \textbf{Reduced object} & \textbf{Compatible cache-level measurements} & \textbf{Benchmark role} \\
\midrule
\endfirsthead
\multicolumn{5}{r}{\small\itshape Table~\thetable\ continued} \\
\toprule
\textbf{Method} & \textbf{Compression decision} & \textbf{Reduced object} & \textbf{Compatible cache-level measurements} & \textbf{Benchmark role} \\
\midrule
\endhead
\midrule
\multicolumn{5}{r}{\small\itshape Continued on next page} \\
\endfoot
\bottomrule
\endlastfoot
\multicolumn{5}{@{}l}{\textbf{Token eviction / position selection}} \\
StreamingLLM~\cite{xiao2024efficient} & Attention sinks plus recency & Token positions & Direct retained-position map & \textbf{Evaluated} \\
H2O~\cite{zhang_h_2o_2023} & Cumulative-attention heavy hitters & Token positions & Direct retained-position map & Survey only \\
Scissorhands~\cite{liu2023scissorhands} & Persistent token importance & Token positions & Direct retained-position map & Survey only \\
TOVA~\cite{oren2024transformers} & Current attention score & Token positions & Direct retained-position map & \textbf{Evaluated} \\
SnapKV~\cite{li_snapkv_2024} & Observation-window pooling & Token clusters / positions & Direct or cluster-projected map & \textbf{Evaluated} \\
KeyDiff~\cite{park2025kediff} & Key-similarity eviction & Token positions & Direct retained-position map & \textbf{Evaluated} \\
\addlinespace[2pt]
\multicolumn{5}{@{}l}{\textbf{Budget allocation / hierarchy}} \\
PyramidKV~\cite{cai2024pyramidkv} & Layer-adaptive funnel & Layer/token quotas & Layer-wise retained-position map & Excluded \\
AdaKV~\cite{feng2024adakv} & Adaptive per-head quota & Head-wise token slots & Direct positions, heads kept separate & \textbf{Evaluated} \\
HeadKV~\cite{fu2024headkv} & Head-level importance and capacity allocation & Head-wise token quotas & Direct positions, heads kept separate & Survey only \\
PolyKV~\cite{fei2026polykv} & Layer-wise policy routing and non-uniform allocation & Layer policies / slot quotas & Per-policy map plus composition metadata & Survey only \\
\addlinespace[2pt]
\multicolumn{5}{@{}l}{\textbf{Query-aware access / channel pruning}} \\
ThinK~\cite{xu2024think} & Query-driven channel salience & Key channels & Structural addressability; ERR/ECov \texttt{N/A}; distribution/decoding & \textbf{Evaluated} \\
Keyformer~\cite{adnan2024keyformer} & Learned key-token score & Token positions & Direct retained-position map & Survey only \\
Quest~\cite{tang2024quest} & Query-aware critical pages & Pages / blocks & Page-to-token projection & Survey only \\
Q-Filters~\cite{godey2025qfilters} & Query--key geometry filter & Tokens / blocks & Query-specific retained-position map & Survey only \\
CHESS~\cite{fei2026chess} & Context-aware hierarchical semantic selection & Grids / chunks / pages & Hierarchy-to-token projection & Survey only \\
\addlinespace[2pt]
\multicolumn{5}{@{}l}{\textbf{Tensor / representation compression}} \\
QuantizedCache (HQQ)~\cite{huggingface_quantizedcache} & Grouped low-bit approximation & Key/value precision & Structural addressability; ERR/ECov \texttt{N/A}; distribution/decoding & \textbf{Evaluated} \\
KIVI~\cite{liu2024kivi} & Asymmetric 2-bit quantization & Key/value precision & Structural addressability plus distribution/decoding metrics & Survey only \\
KVQuant~\cite{hooper2024kvquant} & Non-uniform, outlier-aware quantization & Key/value precision & Structural addressability plus distribution/decoding metrics & Survey only \\
CommVQ~\cite{li2025commvq} & RoPE-commutative vector quantization & Key/value vector codes & Structural addressability plus reconstruction/distribution metrics & Survey only \\
Palu~\cite{chang2025palu} & Low-rank hidden-dimension projection & Latent key/value dimensions & Reconstruction map plus distribution/decoding metrics & Survey only \\
SALS~\cite{mu2025sals} & Sparse attention over a low-rank latent cache & Latent states / selected tokens & Latent reconstruction plus access/distribution metrics & Survey only \\
\addlinespace[2pt]
\multicolumn{5}{@{}l}{\textbf{Chunk / semantic / learned state}} \\
ChunkKV~\cite{liu2025chunkkv} & Semantic chunk selection & Token chunks & Chunk-to-token projection & \textbf{Evaluated} \\
DMC~\cite{nawrot2024dynamic} & Learned state merging & KV states & Merge records plus distribution/decoding measures & Survey only \\
KVzip~\cite{kim2026kvzip} & Context reconstruction & Token/context states & Reconstruction map plus distribution/decoding metrics & Survey only \\
SCOPE~\cite{wu2025scope} & Generation-aware hybrid selection & Tokens / layers & Layer/token map plus distribution/decoding metrics & Survey only \\
\end{longtable}
\endgroup

\noindent\begin{minipage}{\textwidth}
\centering
\small
\setlength{\tabcolsep}{3pt}
\renewcommand{\arraystretch}{1.02}
\captionof{table}{Fixed evaluation configurations. Unless noted, methods use \texttt{kvpress} 0.5.3. The 75/50/25 labels are target retained shares for token-, chunk-, and allocation-based methods; ThinK maps them to key-channel shares, and QuantizedCache maps them to bit widths. They are therefore compression settings, not byte-equivalent cross-method quantities.}
\label{tab:method_configs}
\begin{tabularx}{\textwidth}{@{}
  >{\raggedright\arraybackslash}p{0.14\textwidth}
  >{\raggedright\arraybackslash}p{0.31\textwidth}
  >{\raggedright\arraybackslash}p{0.19\textwidth}
  >{\raggedright\arraybackslash}X@{}}
\toprule
\textbf{Method} & \textbf{Fixed configuration} & \textbf{Compressed unit} & \textbf{Setting interpretation} \\
\midrule
StreamingLLM & \texttt{n\_sink=4} & Token positions & Retained share; four sink tokens remain fixed \\
SnapKV & \texttt{window\_size=64}; \texttt{kernel\_size=5} & Token clusters / positions & Retained share \\
TOVA & No additional fixed parameter & Token positions & Retained share \\
KeyDiff & No additional fixed parameter & Token positions & Retained share \\
ThinK & \texttt{window\_size=32} & Key channels & 75/50 key-channel share; 25 is \texttt{N/A} \\
ChunkKV & K-norm base; \texttt{chunk\_length=20} & Token chunks & Retained share \\
AdaKV & SnapKV base (window 64, kernel 5); \texttt{alpha\_safeguard=0.2} & Head-wise token slots & Base retained share, then budget reallocation \\
QuantizedCache~\cite{huggingface_quantizedcache} & Transformers/HQQ; \texttt{group\_size=64}; \texttt{residual\_length=128} & Key/value precision & 8b/4b/2b; ideal 50/25/12.5\% of bf16 bits \\
\bottomrule
\end{tabularx}
\end{minipage}
\vspace{6pt}

\noindent\begin{minipage}{\textwidth}
\centering
\small
\setlength{\tabcolsep}{4pt}
\captionof{table}{Core execution environment for the reported experiments.}
\label{tab:environment}
\begin{tabular}{@{}p{0.17\textwidth}p{0.29\textwidth}p{0.17\textwidth}p{0.29\textwidth}@{}}
\toprule
\textbf{Field} & \textbf{Recorded value} & \textbf{Field} & \textbf{Recorded value} \\
\midrule
Model ID & \texttt{Qwen/Qwen3-8B} & Exact revision & \makecell[l]{\texttt{b968826d9c46dd6066d109}\\
\texttt{eabc6255188de91218}} \\
PyTorch / CUDA runtime & 2.11.0+cu128 / 12.8 & Transformers / datasets & 5.2.0 / 5.0.0 \\
Accelerate / \texttt{kvpress} & 1.14.0 / 0.5.3 & \texttt{flash-attn} & 2.8.3.post1 \\
Hardware / arithmetic & NVIDIA H200 / bf16 & Quantization / decoder & HQQ / greedy argmax \\
\bottomrule
\end{tabular}
\vspace{6pt}

\parbox{0.90\textwidth}{\footnotesize The public release records the immutable model revision, software environment, method settings, scorer configuration, and artifact checksums.}
\end{minipage}
\vspace{4pt}

\noindent\begin{minipage}{\textwidth}
\centering
\small
\captionof{table}{Verified licenses of the benchmark's source datasets.}
\label{tab:source_licenses}
\begin{tabular*}{0.70\textwidth}{@{\extracolsep{\fill}}ll@{}}
\toprule
\textbf{Source dataset} & \textbf{Verified license} \\
\midrule
RULER & Apache License 2.0 \\
Qasper & CC BY 4.0 \\
HotpotQA & CC BY-SA 4.0 \\
\bottomrule
\end{tabular*}
\vspace{2pt}

\parbox{0.70\textwidth}{\footnotesize The benchmark code and derived artifacts carry explicit compatible licenses and redistribution notices in the public release.}
\end{minipage}
\vspace{5pt}

\subsection{Release Files and Automated Checks}

The release structure specifies what a reusable benchmark contains beyond the paper. Table~\ref{tab:artifact_components} lists the files versioned together and the automated checks that keep them consistent. Each published C$\rightarrow$W row links to its paired FullCache run, diagnostics, and generator script.

\noindent\begin{minipage}{\textwidth}
\centering
\small
\setlength{\tabcolsep}{4pt}
\renewcommand{\arraystretch}{1.05}
\captionof{table}{Release components and automated checks. The components are versioned together so reported rows remain traceable to inputs and generators.}
\label{tab:artifact_components}
\begin{tabularx}{\textwidth}{@{}>{\raggedright\arraybackslash}p{0.17\textwidth}X>{\raggedright\arraybackslash}p{0.24\textwidth}@{}}
\toprule
\textbf{Component} & \textbf{Contents} & \textbf{Automated check} \\
\midrule
Task manifest & Source IDs, references, task identifiers, prompt hashes & Unique IDs; hashes; alignment \\
Core matrix & Workload--method--setting cells, setting interpretation, supported/\texttt{N/A} status & Expected vs. observed rows \\
Paired records & FullCache/compressed outputs, scores, outcome changes, versions & Matching pair; unique key; numeric fields \\
Diagnostic records & Cache, likelihood/distribution, attention applicability, and decoding measurements & Required fields present; values finite and nonempty \\
Validation summaries & Task-score and C$\rightarrow$W CSVs, overlap sets, figure inputs & Regeneration; counts sum to totals \\
Reproduction package & Fixed dependency versions, pre-run report, generator scripts & Version checks; deterministic rerun \\
\bottomrule
\end{tabularx}
\end{minipage}

\subsection{Results on the Full Evaluation Sets}

Table~\ref{tab:layer1_ruler} reports task quality and C$\rightarrow$W rates for every supported method--setting cell in the four fixed evaluation sets.

\begin{table}[!ht]
\centering
\fontsize{8.2pt}{9pt}\selectfont
\setlength{\tabcolsep}{3pt}
\renewcommand{\arraystretch}{1.01}
\caption{Results for all sources across the four workloads. Compressed cells report task score / C$\rightarrow$W rate (\%); the FullCache row reports task / binary score and the FullCache-correct denominator $n_{\mathrm{FC}}$.}
\label{tab:layer1_ruler}
\label{tab:layer1_qa}
\begin{tabular*}{\textwidth}{@{\extracolsep{\fill}}lcrrrr@{}}
\toprule
\textbf{Method} & \textbf{Setting} & \textbf{RULER-8K} & \textbf{RULER-16K} & \textbf{Qasper} & \textbf{HotpotQA} \\
\midrule
\multicolumn{2}{l}{\textbf{FullCache task / binary}} & \makecell{97.1 / 88.0\\$n_{\mathrm{FC}}=968$} & \makecell{96.8 / 86.5\\$n_{\mathrm{FC}}=951$} & \makecell{29.5 / 29.5\\$n_{\mathrm{FC}}=59$} & \makecell{74.0 / 74.0\\$n_{\mathrm{FC}}=148$} \\
\midrule
\multirow{3}{*}{StreamingLLM}
 & 75\% & 79.7 / 29.6 & 76.8 / 31.8 & 25.0 / 20.3 & 68.0 / 10.1 \\
 & 50\% & 57.2 / 56.5 & 54.9 / 58.4 & 21.5 / 33.9 & 57.5 / 24.3 \\
 & 25\% & 32.9 / 81.5 & 33.6 / 77.6 & 17.5 / 44.1 & 50.0 / 33.8 \\
\midrule
\multirow{3}{*}{SnapKV}
 & 75\% & 96.6 / 2.5  & 96.8 / 1.4  & 31.5 / 1.7  & 71.5 / 3.4 \\
 & 50\% & 90.8 / 11.5 & 95.3 / 3.6  & 27.5 / 8.5  & 69.0 / 6.8 \\
 & 25\% & 82.9 / 25.4 & 86.3 / 16.8 & 27.5 / 11.9 & 63.0 / 14.9 \\
\midrule
\multirow{3}{*}{TOVA}
 & 75\% & 96.9 / 1.2  & 96.7 / 0.9  & 30.0 / 0.0  & 72.5 / 2.0 \\
 & 50\% & 95.2 / 5.0  & 95.5 / 3.5  & 29.0 / 10.2 & 70.0 / 5.4 \\
 & 25\% & 84.6 / 22.0 & 85.5 / 17.5 & 27.5 / 16.9 & 65.5 / 11.5 \\
\midrule
\multirow{3}{*}{KeyDiff}
 & 75\% & 96.6 / 5.4  & 96.3 / 4.1  & 24.5 / 18.6 & 64.5 / 16.2 \\
 & 50\% & 90.7 / 14.5 & 90.9 / 12.0 & 18.0 / 40.7 & 51.5 / 34.5 \\
 & 25\% & 78.6 / 30.3 & 78.9 / 28.1 & 13.5 / 57.6 & 25.5 / 66.2 \\
\midrule
\multirow{3}{*}{ThinK}
 & 75\% & 97.7 / 2.2  & 97.2 / 1.8  & 28.0 / 10.2 & 68.5 / 10.8 \\
 & 50\% & 0.0 / 100.0 & 0.0 / 100.0 & 6.5 / 78.0  & 12.5 / 83.8 \\
 & 25\% & \texttt{N/A} & \texttt{N/A} & \texttt{N/A} & \texttt{N/A} \\
\midrule
\multirow{3}{*}{ChunkKV}
 & 75\% & 89.9 / 17.7 & 89.4 / 17.2 & 24.0 / 23.7 & 60.5 / 23.6 \\
 & 50\% & 76.6 / 37.1 & 78.3 / 32.6 & 21.0 / 37.3 & 45.0 / 43.9 \\
 & 25\% & 58.5 / 53.0 & 62.4 / 47.1 & 18.5 / 40.7 & 28.5 / 66.9 \\
\midrule
\multirow{3}{*}{AdaKV}
 & 75\% & 96.9 / 1.5  & 96.8 / 0.7  & 30.5 / 0.0  & 71.5 / 3.4 \\
 & 50\% & 93.4 / 7.3  & 96.3 / 2.4  & 27.5 / 10.2 & 70.0 / 5.4 \\
 & 25\% & 84.1 / 24.1 & 88.3 / 13.8 & 27.0 / 11.9 & 65.5 / 12.2 \\
\midrule
\multirow{3}{*}{QuantizedCache}
 & 8b & 97.0 / 0.9  & 96.9 / 0.4  & 29.5 / 1.7  & 72.5 / 2.0 \\
 & 4b & 97.1 / 2.8  & 96.0 / 2.9  & 29.5 / 5.1  & 70.0 / 5.4 \\
 & 2b & 20.0 / 84.8 & 9.2 / 92.8  & 8.5 / 71.2  & 21.0 / 72.3 \\
\bottomrule
\end{tabular*}
\vspace{2pt}

\parbox{\textwidth}{\footnotesize Values are exact for all sources in the fixed evaluation sets. Settings order compression severity within a method and are not byte-equivalent. Qasper and HotpotQA have only 59 and 148 FullCache-correct instances, respectively; their values describe this evaluation set.}
\end{table}
\FloatBarrier

\section{Release Validation and Disclosure}

\begin{table}[!t]
\centering
\small
\setlength{\tabcolsep}{4pt}
\renewcommand{\arraystretch}{1.04}
\caption{Additional RULER-8K diagnostic-corpus coverage beyond the aggregate run accounting in Table~\ref{tab:completed_run_accounting}.}
\label{tab:app_ruler8k_accounting}
\begin{tabular}{@{}p{0.27\textwidth}p{0.60\textwidth}r@{}}
\toprule
\textbf{Item} & \textbf{Definition} & \textbf{Count} \\
\midrule
Aggressive tier & 25\% retention or 2b quantization & 3,108 \\
Intermediate tier & 50\% retention or 4b quantization & 2,271 \\
Light tier & 75\% retention or 8b quantization & 591 \\
Coverage applicability status & Measured, projected, structural, or \texttt{N/A} recorded & 5,970 / 5,970 \\
Numeric ERR/ECov rows & Valid direct/projected original-token maps & 4,124 / 5,970 \\
EAR rows & EAR computed from valid eager-attention traces; other rows carry explicit \texttt{N/A} & 4,608 / 5,970 \\
Auxiliary demand labels & Five context-demand labels available & 5,970 / 5,970 \\
\midrule
Most frequent method & StreamingLLM selected rows & 1,623 \\
Least frequent method & TOVA selected rows & 273 \\
\bottomrule
\end{tabular}
\end{table}

\paragraph{AI-use disclosure.}
This paper was prepared with AI-assisted tools for literature-search strategy, structure planning, drafting, citation checks, simulated review, and formatting. AI assistance also informed the auxiliary context-demand criteria and lookup generator; final labels are assigned deterministically without per-sample model inference. All arguments, claims, and conclusions were directed and reviewed by the authors, who take responsibility for the work.

Adapter tests verify that each specialized \texttt{compress} implementation invokes the intended method. A CPU-only command regenerates run counts, tables, and figures, and a small GPU run repeats selected pairs. The package records artifact checksums, compatible licenses, a correction channel, and a permanent release URL.

\begin{figure}[t]
  \centering
  \includegraphics[width=.8\textwidth]{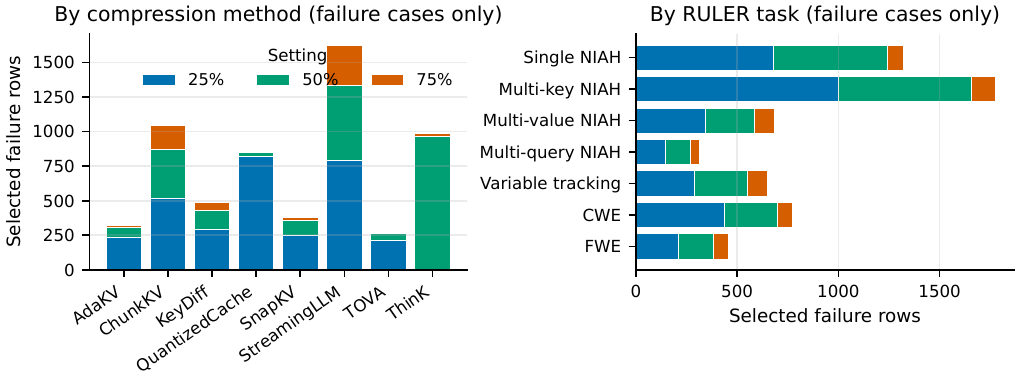}
  \caption{Composition of the corrected RULER-8K diagnostic corpus. Left: selected failure rows by setting tier; 25/50/75\% correspond to 2b/4b/8b for QuantizedCache. Right: selected rows by RULER task family.}
  \Description{Two bar-chart panels show selected failure counts by eight compression methods, compression setting, and RULER task family.}
  \label{fig:app_ruler8k_corpus_composition}
\end{figure}

Figure~\ref{fig:app_ruler8k_corpus_composition} complements Table~\ref{tab:app_ruler8k_accounting} by showing how the corrected RULER-8K failure rows are distributed across method--setting tiers and task families. It is a composition view of the selected corpus, not a ranking of method or task difficulty.

\begin{table}[t]
\centering
\fontsize{8.5pt}{9.4pt}\selectfont
\setlength{\tabcolsep}{2pt}
\renewcommand{\arraystretch}{1.04}
\caption{RULER-8K task composition and demand profile. ERR/ECov average only measured/projected rows; other metric columns use their own valid rows. EAR and access demand exclude AdaKV/ChunkKV rows without eager-attention traces.}
\label{tab:app_ruler8k_task_profile}
\begin{tabular*}{\textwidth}{@{\extracolsep{\fill}}lrrrrrrrrrr@{}}
\toprule
\textbf{Task} & \textbf{Rows} & \textbf{Sources} & \makecell{\textbf{Slot}\\\textbf{ERR}} & \makecell{\textbf{Slot}\\\textbf{ECov}} & \textbf{$\Delta$NLL} & \textbf{Top-50} & \textbf{EAR} & \textbf{Evidence} & \makecell{\textbf{Distribution}\\\textbf{demand}} & \makecell{\textbf{Access}\\\textbf{demand}} \\
\midrule
Single NIAH & 1,319 & 300 & 0.212 & 0.234 & 3.311 & 0.587 & 0.120 & 1.000 & 1.996 & 2.311 \\
Multi-key NIAH & 1,777 & 300 & 0.207 & 0.151 & 2.766 & 0.598 & 0.085 & 2.000 & 2.001 & 2.248 \\
Multi-value NIAH & 682 & 95 & 0.559 & 0.589 & 1.342 & 0.798 & 0.231 & 2.000 & 0.926 & 1.355 \\
Multi-query NIAH & 314 & 39 & 0.593 & 0.626 & 1.458 & 0.774 & 0.225 & 2.000 & 1.001 & 1.249 \\
Variable tracking & 652 & 100 & 0.519 & 0.520 & 4.509 & 0.572 & 0.309 & 3.000 & 2.096 & 1.220 \\
CWE & 771 & 72 & 0.436 & 0.444 & 1.228 & 0.884 & 0.222 & 3.000 & 0.591 & 1.501 \\
FWE & 455 & 62 & 0.705 & 0.713 & 1.303 & 0.819 & 0.303 & 3.000 & 0.577 & 1.541 \\
\bottomrule
\end{tabular*}
\end{table}

\section{RULER-8K Paired Failure Analysis}
\label{app:ruler8k_complete}

This section reports task composition, row-level cases, and method--setting diagnostics for the RULER-8K C$\rightarrow$W rows defined in Section~\ref{sec:all_source_evaluation}. Metric definitions and applicability conditions follow Section~\ref{sec:metric}; a source may contribute multiple failure rows.

\subsection{Diagnostic Corpus Composition}

Single-needle and multi-key retrieval dominate the failure subset and have high predictive-distribution and attention-access demand. Variable tracking has the largest mean $\Delta$NLL in Table~\ref{tab:app_ruler8k_task_profile}; retaining the relevant positions is insufficient when compression disrupts the model state needed to associate variables with their values. Word extraction has high evidence demand but lower predictive-distribution demand, making it useful for separating evidence coverage from formatting and extraction accuracy.

\subsection{Diagnostics by Method and Setting}

\begin{table}[t]
\centering
\small
\setlength{\tabcolsep}{2.8pt}
\renewcommand{\arraystretch}{1.02}
\caption{Exact method-by-setting diagnostics on completed RULER-8K C$\rightarrow$W rows. ERR/ECov are measured except for projected ChunkKV; $S$ marks structural position addressability, for which both are \texttt{N/A}. $\dagger$ marks unavailable native EAR. Demand is predictive-distribution/access (D/A). Rows/src. reports failure rows/unique sources.}
\label{tab:app_ruler8k_method_budget}
\begin{tabular*}{\textwidth}{@{\extracolsep{\fill}}lcrrrrrrrr@{}}
\toprule
\textbf{Method} & \textbf{Setting} & \makecell{\textbf{Rows/}\\\textbf{src.}} & \textbf{ERR} & \textbf{ECov} & \textbf{$\Delta$NLL} & \textbf{KL} & \textbf{Top-50} & \textbf{EAR} & \textbf{Demand D/A} \\
\midrule
\multirow{3}{*}{AdaKV}
 & 25\% & 233/233 & 0.454 & 0.486 & 0.795 & 0.779 & 0.759 & \texttt{N/A}$^\dagger$ & 0.870/\texttt{N/A} \\
 & 50\% & 71/71   & 0.645 & 0.686 & 0.202 & 0.202 & 0.837 & \texttt{N/A}$^\dagger$ & 0.545/\texttt{N/A} \\
 & 75\% & 15/15   & 0.868 & 0.880 & 0.003 & 0.001 & 0.970 & \texttt{N/A}$^\dagger$ & 0.044/\texttt{N/A} \\
\midrule
\multirow{3}{*}{ChunkKV}
 & 25\% & 513/513 & 0.430 & 0.424 & 1.633 & 1.623 & 0.711 & \texttt{N/A}$^\dagger$ & 1.353/\texttt{N/A} \\
 & 50\% & 359/359 & 0.578 & 0.549 & 1.143 & 1.154 & 0.759 & \texttt{N/A}$^\dagger$ & 0.965/\texttt{N/A} \\
 & 75\% & 171/171 & 0.702 & 0.693 & 0.696 & 0.699 & 0.803 & \texttt{N/A}$^\dagger$ & 0.672/\texttt{N/A} \\
\midrule
\multirow{3}{*}{KeyDiff}
 & 25\% & 293/293 & 0.437 & 0.356 & 1.043 & 1.063 & 0.744 & 0.076 & 1.039/2.075 \\
 & 50\% & 140/140 & 0.673 & 0.653 & 0.531 & 0.513 & 0.799 & 0.084 & 0.674/1.907 \\
 & 75\% & 52/52   & 0.885 & 0.952 & 0.097 & 0.099 & 0.886 & 0.101 & 0.263/1.577 \\
\midrule
\multirow{3}{*}{Quant.}
 & 2b & 821/821 & \texttt{N/A}$^S$ & \texttt{N/A}$^S$ & 2.404 & 2.397 & 0.581 & 0.554 & 1.948/0.730 \\
 & 4b & 27/27   & \texttt{N/A}$^S$ & \texttt{N/A}$^S$ & 0.007 & 0.003 & 0.947 & 0.969 & 0.037/0.000 \\
 & 8b & 9/9     & \texttt{N/A}$^S$ & \texttt{N/A}$^S$ & 0.000 & 0.000 & 0.989 & 0.998 & 0.000/0.000 \\
\midrule
\multirow{3}{*}{SnapKV}
 & 25\% & 246/246 & 0.465 & 0.488 & 0.897 & 0.886 & 0.760 & 0.047 & 0.932/2.028 \\
 & 50\% & 111/111 & 0.641 & 0.682 & 0.345 & 0.358 & 0.807 & 0.047 & 0.691/1.982 \\
 & 75\% & 24/24   & 0.870 & 0.887 & 0.033 & 0.037 & 0.954 & 0.052 & 0.125/2.000 \\
\midrule
\multirow{3}{*}{Stream.}
 & 25\% & 789/789 & 0.079 & 0.079 & 2.133 & 2.151 & 0.681 & 0.046 & 1.676/2.548 \\
 & 50\% & 547/547 & 0.172 & 0.171 & 1.955 & 1.977 & 0.699 & 0.074 & 1.544/2.333 \\
 & 75\% & 287/287 & 0.258 & 0.258 & 1.506 & 1.534 & 0.744 & 0.087 & 1.273/2.206 \\
\midrule
\multirow{3}{*}{TOVA}
 & 25\% & 213/213 & 0.467 & 0.454 & 0.833 & 0.783 & 0.766 & 0.159 & 0.892/1.385 \\
 & 50\% & 48/48   & 0.707 & 0.716 & 0.177 & 0.191 & 0.874 & 0.196 & 0.410/1.125 \\
 & 75\% & 12/12   & 0.937 & 0.956 & 0.072 & 0.075 & 0.969 & 0.228 & 0.083/1.000 \\
\midrule
\multirow{3}{*}{ThinK}
 & 25\% & \texttt{N/A} & \texttt{N/A} & \texttt{N/A} & \texttt{N/A} & \texttt{N/A} & \texttt{N/A} & \texttt{N/A} & \texttt{N/A} \\
 & 50\% & 968/968 & \texttt{N/A}$^S$ & \texttt{N/A}$^S$ & 7.827 & 7.821 & 0.486 & 0.063 & 2.847/2.148 \\
 & 75\% & 21/21   & \texttt{N/A}$^S$ & \texttt{N/A}$^S$ & 0.013 & 0.009 & 0.926 & 1.012 & 0.063/0.000 \\
\bottomrule
\end{tabular*}
\end{table}

\begin{figure*}[!t]
  \centering
  \includegraphics[width=0.9\textwidth]{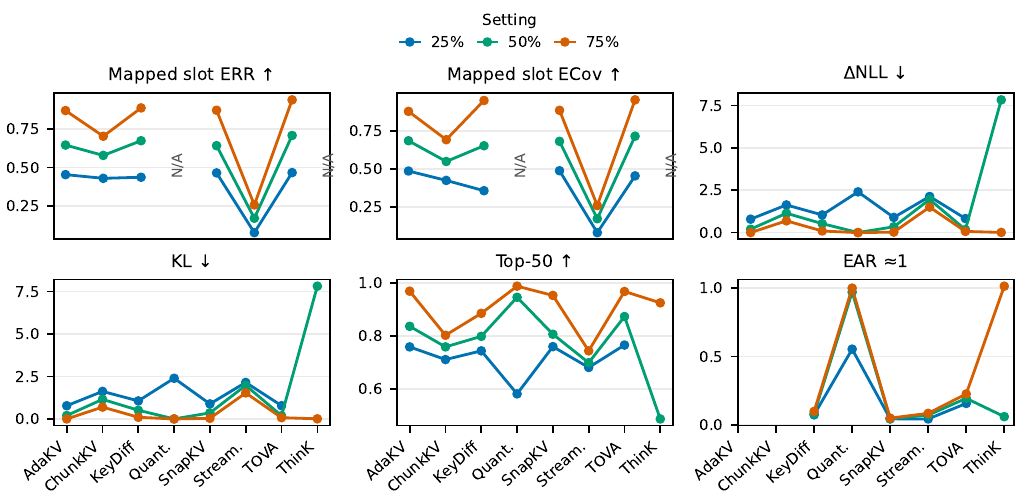}
\caption{Visual summary of Table~\ref{tab:app_ruler8k_method_budget}. Structural ERR/ECov cells are \texttt{N/A}, not one. Higher is preferred for applicable ERR, ECov, and Top-50; lower is preferred for $\Delta$NLL and KL. AdaKV and ChunkKV have no valid native EAR.}
  \Description{Six compact panels show ERR, ECov, delta NLL, KL, Top-50, and valid EAR by eight methods and three compression settings.}
  \label{fig:app_ruler8k_metric_profiles}
\end{figure*}

Means in Table~\ref{tab:app_ruler8k_method_budget} are conditional on each method's own C$\rightarrow$W set; direct numerical comparisons require common-failure intersections. Figure~\ref{fig:app_ruler8k_metric_profiles} visualizes the corresponding method--setting profiles while preserving structural and unavailable measurements as \texttt{N/A}.

\begin{table}[t]
\centering
\small
\setlength{\tabcolsep}{3pt}
\renewcommand{\arraystretch}{1.06}
\caption{RULER-8K row-level examples under the corrected slot estimator. Stable local predictive distributions, low EAR, and a wrong free-generation output suggest different explanations; they are not causal labels.}
\label{tab:app_ruler8k_case_studies}
\begin{tabular}{@{}
  >{\raggedright\arraybackslash}p{0.16\textwidth}
  >{\raggedright\arraybackslash}p{0.21\textwidth}
  >{\raggedright\arraybackslash}p{0.33\textwidth}
  >{\raggedright\arraybackslash}p{0.22\textwidth}@{}}
\toprule
\textbf{Case} & \textbf{Run} & \textbf{Paired output and diagnostics} & \textbf{Interpretation} \\
\midrule
Long-range low coverage
& \texttt{vt\_000097}\newline StreamingLLM / 25\%
& FC: QDVHI; Comp.: Q, R, S, T, U\newline
  ECov 0.000; $\Delta$NLL 7.033; Top-50 0.420; EAR 0.000
& Support disappears across slots; selection is the first component to examine. \\
Retained but low EAR
& \texttt{niah\_multikey\_3\_004079}\newline KeyDiff / 75\%
& FC: \texttt{e2e94afd-7900-42e3-}\newline
  \texttt{8a73-7e8c19d14bf2}\newline
  Comp.: \texttt{e2e94afd-7900-42e3-}\newline
  \texttt{8a73-7e8c1d9e4bf2.}\newline
  ECov 0.958; $\Delta$NLL 0.137; KL 0.288; Top-50 0.706; EAR 0.015
& Coverage is high while EAR is near zero; access is the only abnormal diagnostic. \\
Conflicting access/decoding evidence
& \texttt{cwe\_000045}\newline ThinK / 50\%
& FC: massage; Comp.: $\varnothing$\newline
  ECov \texttt{N/A}$^S$; $\Delta$NLL 0.000; KL 0.000; Top-50 1.000; EAR 0.228
& KL/Top-50 are stable, EAR is low, and free generation is empty; attribution is not unique. \\
Decoding or scoring candidate
& \texttt{cwe\_000004}\newline QuantizedCache / 2b
& FC: vendor; Comp.: subsidy\newline
  ECov \texttt{N/A}$^S$; $\Delta$NLL 0.000; KL 0.000; Top-50 1.000; EAR 0.787
& Positions are structurally addressable and local distribution/EAR are stable, but the emitted list fails the scorer. \\
\bottomrule
\end{tabular}
\end{table}

The cases in Table~\ref{tab:app_ruler8k_case_studies} apply the metrics to concrete rows. The StreamingLLM case removes both mapped evidence and attention access. The KeyDiff case retains 95.8\% slot coverage but has EAR 0.015 and a corrupted identifier. ThinK and QuantizedCache show continuation failure despite structural position addressability; neither has a numeric coverage claim. The final two rows separate a conflicting empty-output case from a decoding/scoring candidate with stable local measurements; neither observation alone identifies the cause.

\subsection{Auxiliary Demand Annotations and Profiles}

The deterministic task-template labels and their aggregate distribution are defined in Section~\ref{sec:analysis}. The figure below retains the fuller distributional view used for secondary analysis.

\begin{figure}[t]
  \centering
  \includegraphics[width=.45\columnwidth]{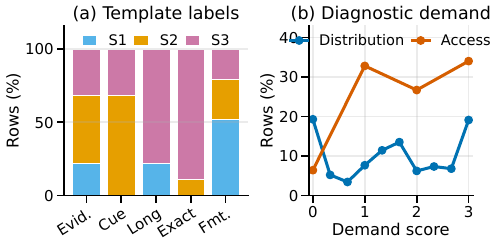}
  \caption{Deterministic task-template label distributions (left) and metric-derived demand scores (right) on corrected RULER-8K failure rows. Access demand uses only rows with valid EAR.}
  \Description{Side-by-side bar-chart and line-plot panels show five auxiliary context-demand axes and distribution/access demand-score distributions.}
  \label{fig:app_ruler8k_demand_distributions}
\end{figure}

Figure~\ref{fig:app_ruler8k_demand_distributions} reports the auxiliary labels together with metric-derived demand-score distributions. These profiles condition on observed failures and do not imply general task difficulty or a method ranking.

\FloatBarrier

\section{RULER-16K and Evidence-Annotated QA Results}
\label{app:completed_extra_results}

For RULER-16K, the light tier (75\% retention or 8b) contains 555 failures among 7,608 FullCache-correct supported pairs, the intermediate tier (50\% or 4b) contains 2,048 among 7,608, and the aggressive tier (25\% or 2b) contains 2,793 among 6,657 because ThinK is unsupported. The table reports diagnostics conditional on each method's own failures.

For evidence-annotated QA, the same tiers contain 151, 442, and 561 C$\rightarrow$W rows, respectively: 45/132/150 on Qasper and 106/310/411 on HotpotQA. These counts sum to the 327 and 827 rows in Table~\ref{tab:completed_run_accounting}; Figure~\ref{fig:qa_transfer} reports their diagnostic-category composition. Evidence is aligned after final-prompt tokenization as described in Section~\ref{sec:task_method_adapters}, and an unavailable mapping makes only the affected cache diagnostic \texttt{N/A}, not the source or paired outcome.

\noindent\begin{minipage}{\textwidth}
\centering
\footnotesize
\setlength{\tabcolsep}{2.4pt}
\renewcommand{\arraystretch}{0.98}
\captionof{table}{RULER-16K diagnostics on each method's own C$\rightarrow$W rows. ECov is measured except for projected ChunkKV; $S$ marks structural addressability with ECov \texttt{N/A}. The category is the modal row-level label before aggregation.}
\label{tab:ruler16k_failure_metrics}
\begin{tabular}{@{}lcrrrrp{0.25\textwidth}@{}}
\toprule
\textbf{Method} & \textbf{Setting} & \makecell{\textbf{C$\rightarrow$W}\\\textbf{rows}} & \textbf{ECov} & \textbf{$\Delta$NLL} & \textbf{GPR} & \makecell[l]{\textbf{Most frequent row-level}\\\textbf{diagnostic category}} \\
\midrule
\textsc{StreamingLLM}~\cite{xiao2024efficient} & 75\% & 302 & 0.260 & 1.689 & 0.480 & Low mapped coverage \\
\textsc{SnapKV}~\cite{li_snapkv_2024} & 75\% & 13 & 0.812 & 0.038 & 0.970 & Partial mapped coverage \\
\textsc{TOVA}~\cite{oren2024transformers} & 75\% & 9 & 0.932 & 0.000 & 1.000 & Conflicting diagnostic signals \\
\textsc{KeyDiff}~\cite{park2025kediff} & 75\% & 39 & 0.944 & 0.105 & 0.920 & Conflicting diagnostic signals \\
\textsc{ThinK}~\cite{xu2024think} & 75\% & 17 & \texttt{N/A}$^S$ & 0.023 & 0.983 & Decoding/scoring candidate \\
\textsc{ChunkKV}~\cite{liu2025chunkkv} & 75\% & 164 & 0.698 & 0.423 & 0.835 & Partial mapped coverage \\
\textsc{AdaKV}~\cite{feng2024adakv} & 75\% & 7 & 0.805 & 0.000 & 1.000 & Partial mapped coverage \\
\textsc{QuantizedCache}~\cite{huggingface_quantizedcache} & 8b & 4 & \texttt{N/A}$^S$ & 0.037 & 0.965 & Decoding/scoring candidate \\
\midrule
\textsc{StreamingLLM} & 50\% & 555 & 0.160 & 2.107 & 0.341 & Low mapped coverage \\
\textsc{SnapKV} & 50\% & 34 & 0.639 & 0.117 & 0.908 & Partial mapped coverage \\
\textsc{TOVA} & 50\% & 33 & 0.665 & 0.135 & 0.900 & Partial mapped coverage \\
\textsc{KeyDiff} & 50\% & 114 & 0.594 & 0.595 & 0.681 & Low mapped coverage \\
\textsc{ThinK} & 50\% & 951 & \texttt{N/A}$^S$ & 7.958 & 0.015 & Structural-position drift \\
\textsc{ChunkKV} & 50\% & 310 & 0.567 & 0.870 & 0.643 & Partial mapped coverage \\
\textsc{AdaKV} & 50\% & 23 & 0.584 & -0.017 & 1.034 & Low mapped coverage \\
\textsc{QuantizedCache} & 4b & 28 & \texttt{N/A}$^S$ & 0.002 & 0.998 & Decoding/scoring candidate \\
\midrule
\textsc{StreamingLLM} & 25\% & 738 & 0.070 & 2.290 & 0.260 & Low mapped coverage \\
\textsc{SnapKV} & 25\% & 160 & 0.488 & 0.502 & 0.684 & Low mapped coverage \\
\textsc{TOVA} & 25\% & 166 & 0.445 & 0.759 & 0.601 & Low mapped coverage \\
\textsc{KeyDiff} & 25\% & 267 & 0.353 & 1.065 & 0.558 & Low mapped coverage \\
\textsc{ThinK} & 25\% & \texttt{N/A} & \texttt{N/A} & \texttt{N/A} & \texttt{N/A} & Unsupported \\
\textsc{ChunkKV} & 25\% & 448 & 0.467 & 1.395 & 0.467 & Low mapped coverage \\
\textsc{AdaKV} & 25\% & 131 & 0.484 & 0.409 & 0.724 & Low mapped coverage \\
\textsc{QuantizedCache} & 2b & 883 & \texttt{N/A}$^S$ & 2.230 & 0.260 & Structural-position drift \\
\bottomrule
\end{tabular}
\end{minipage}